\documentclass{article}

\usepackage{microtype}
\usepackage{graphicx}
\usepackage{subfigure}
\usepackage{booktabs} 

\usepackage{hyperref}

\usepackage[accepted]{icml2024}

\usepackage{amsmath}
\usepackage{amssymb}
\usepackage{mathtools}
\usepackage{amsthm}
\usepackage{xspace}
\usepackage{commath}
\usepackage{spverbatim}

\usepackage{lscape}
\usepackage{longtable}

\usepackage[capitalize,noabbrev]{cleveref}

\theoremstyle{plain}
\newtheorem{theorem}{Theorem}[section]
\newtheorem{proposition}[theorem]{Proposition}

\theoremstyle{definition}

\theoremstyle{remark}

\newcommand{\ourmethod}{VAS\xspace}

\usepackage[textsize=tiny]{todonotes}

\begin{document}

\twocolumn[
\icmltitle{Value Augmented Sampling for Language Model Alignment and Personalization}



\icmlsetsymbol{equal}{*}

\begin{icmlauthorlist}
\icmlauthor{Seungwook Han}{equal,IAI,MIT}
\icmlauthor{Idan Shenfeld}{equal,IAI,MIT}
\icmlauthor{Akash Srivastava}{IBM}
\icmlauthor{Yoon Kim}{MIT}
\icmlauthor{Pulkit Agrawal}{IAI,MIT}
\end{icmlauthorlist}

\icmlaffiliation{IAI}{Improbable AI Lab}
\icmlaffiliation{MIT}{MIT}
\icmlaffiliation{IBM}{MIT-IBM Watson AI Labs}

\icmlcorrespondingauthor{Idan Shenfeld}{idanshen@mit.edu}
\icmlcorrespondingauthor{Seungwook Han}{swhan@mit.edu}

\icmlkeywords{Machine Learning, ICML}

\vskip 0.3in
]



\printAffiliationsAndNotice{\icmlEqualContribution} 

\begin{abstract}
Aligning Large Language Models (LLMs) to cater to different human preferences, learning new skills, and unlearning harmful behavior is an important problem. Search-based methods, such as Best-of-N or Monte-Carlo Tree Search, are performant, but impractical for LLM adaptation due to their high inference cost. 
On the other hand, using Reinforcement Learning (RL) for adaptation is computationally efficient, but performs worse due to the optimization challenges in co-training the value function and the policy. We present a new framework for reward optimization, \textit{Value Augmented Sampling} (\ourmethod), that can maximize different reward functions using data sampled from only the initial, frozen LLM. \ourmethod solves for the optimal reward-maximizing policy without co-training the policy and the value function, making the optimization stable, outperforming established baselines, such as PPO and DPO, on standard benchmarks, and achieving comparable results to Best-of-128 with lower inference cost. 
Unlike existing RL methods that require changing the weights of the LLM, \ourmethod does not require access to the weights of the pre-trained LLM. Thus, it can even adapt LLMs (e.g., ChatGPT), which are available only as APIs.
In addition, our algorithm unlocks the new capability of composing several rewards and controlling the extent of each one during deployment time, paving the road ahead for the future of aligned, personalized LLMs.

\end{abstract}

\section{Introduction}
Large Language Models (LLMs) trained on large amounts of Internet data are useful for many applications, such as programming and web-based search \cite{schick2023toolformer, rozière2024code, vicuna2023}. It is common practice to adapt pre-trained language models to improve their task-specific capabilities, suppress unwanted behavior, or change their response style. 
A popular approach for adaptation is to represent a target task or behavior as a reward function and leverage one of many search/learning-based frameworks to maximize the reward. 

Different reward optimization methods have their strengths and weaknesses. In our opinion, there are \textit{five} essential characteristics: 
(i) \textit{High Performance}: The ability to effectively maximize the specified reward, while maintaining other desirable capabilities of the LLM \cite{askell2021general}; 
(ii) \textit{Computational Efficiency} during inference is critical for real-world deployment; 
(iii) \textit{Fine-Grained Adaptation}: The ability to easily tune the extent of reward optimization during deployment according to the user's preference. For example, providing the user with a knob to specify how much alignment should be performed for a particular reward function, such as conciseness; 
(iv) \textit{On-the-fly Composition}: The ability to align the LLM to new, multiple reward functions during deployment (i.e., on-the-fly) without modifying the initial LLM. Such ability will enable users to express new preferences continually without the computational burden of re-training the LLM to be adapted to the combination of existing and new preferences; 
(v) \textit{Black-Box Adaptation}: 
As many state-of-the-art models are available only via API \cite{openai2024gpt4, reid2024gemini}, we desire the ability to adapt both white and black-box models.

A straightforward way to maximize the reward is to generate $N$ different responses and choose the one with the highest reward. This method, called Best-of-N (BoN), achieves high performance with a sufficiently large $N$ \cite{gao2023scaling}. Moreover, since the search is conducted during deployment, BoN can be used for on-the-fly composition, adapting black-box models, and, in principle, fine-grained adaptation. 
Despite their superior performance and other advantages, BoN and other search-based methods \citep{silver2017mastering} are not used for real-world deployment due to their high computational expense (e.g., BoN requires $N$ inferences from the LLM).

To mitigate the high inference-time cost, it is common practice to modify the weights of the initial LLM to directly generate the reward-maximizing response. In this view, the LLM can be considered as the ``policy", and actor-critic reinforcement learning (RL), such as Proximal Policy Optimization (PPO)~\citep{schulman2017proximal, wu2023pairwise} has emerged as the prominent method for learning the policy weights. 
While such adaptation is computationally efficient to deploy, it lags behind the naive BoN search in terms of performance \citep{gao2023scaling}. 

We hypothesize that actor-critic RL performs worse than BoN because of challenges posed by bi-level optimization alternating between learning an estimator of the expected reward (i.e., the value function) of the current policy and improving the policy using the estimated value function. We conjecture that such circular dependence results in the convergence of actor-critic methods to sub-optimal solutions
because (1) training a value estimator for a non-stationary policy is known to be a challenge \citep{nikishin2022primacy, lyle2022understanding}, (2) errors in value estimation can result in noisy policy updates  \cite{fujimoto2018addressing}
, and (3) in LLM adaptation, it is common to constrain the learned policy to be close to the original LLM to preserve the LLM's capabilities, such as reasoning \cite{2309.06256}.
Solving constrained optimization problems with actor-critic RL is a well-known challenge \citep{liu2021policy}.
In addition to the performance gap, learning-based 
methods require access to the models' weights and cannot perform on-the-fly composition or fine-grained adaptation during deployment, which is necessary for user personalization. 

In this paper, we present a reward-maximization method that aims to attain the five essential characteristics of LLM adaptation outlined above. Our key insight is that under the common assumption that the pre-trained LLM only needs to be modified by a small amount, alternating training of the policy and value function is unnecessary. Instead, we estimate the value function of the reward function using data collected from the base (i.e., pre-trained) LLM policy. Thus, we bypass the bi-level optimization process, which is the cause of sub-optimality in actor-critic RL methods. 
Furthermore, unlike BoN, where the search is performed by evaluating entire sequences, we guide the search token-by-token using the value function as the search heuristic. Thereby, we achieve superior performance given the same computational budget. Finally, our formulation, which keeps the weights of the base LLM frozen, unlocks new capabilities -- the ability to adapt LLMs to new and multiple user preferences during deployment without the need for any LLM re-training and adapting black-box LLMs whose weights are unavailable. 

In summary, Value Augmented Sampling (\ourmethod):
\begin{enumerate}
    \item Overcomes the instabilities of existing RL algorithms for LLM adaptation and consequently outperforms strong baselines, such as PPO and DPO \cite{rafailov2023direct} on the benchmark tasks of improving summarization \cite{clark2023seahorse} and multi-turn chat dialogue \cite{bai2022training}. 
    \item\ourmethod is on par with Best-of-128 in both tasks while being at least x6 more computationally efficient in terms of FLOPS.
    \item \ourmethod offers the ability to perform fine-grained control over the magnitude at which the model is being adapted towards a particular reward at inference time. We show how this capability can be used to personalize the LLM's output for a specific user preference.
    \item \ourmethod can adapt closed-source, proprietary models with only API access. We demonstrate that we can improve GPT-3.5 capability for new tasks without access to the model’s weights.
\end{enumerate}

\begin{figure*}[t]
    \centering
    \includegraphics[width=0.85\linewidth]{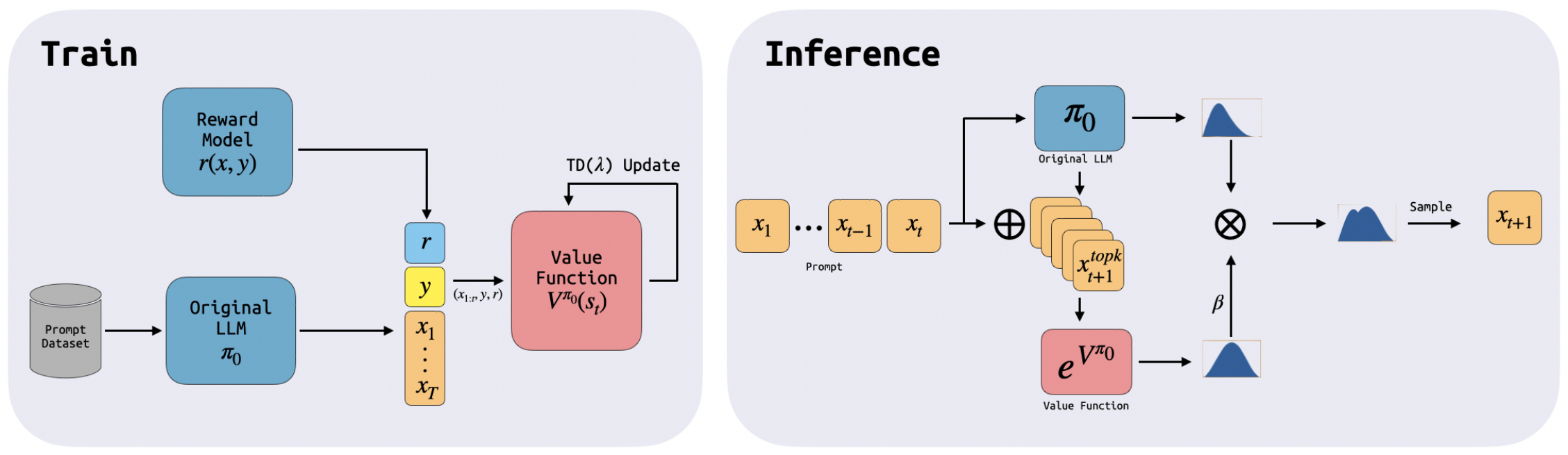}
    \vspace*{-3mm}
    \caption{Overview of \ourmethod training and inference. During training, our algorithm learns a Value function estimator using TD learning. In inference, we augment the output distribution of the original LLM, $\pi_0$, with the exponentiated Values scaled by a weight coefficient $\beta$.}
    \label{fig:algo}
    \vspace*{-3mm}
\end{figure*}

\section{Related Work}
\label{sec:related}
\paragraph{Search-based algorithms for LLM decoding.} To maximize the reward, theoretically, one could conduct a full search of all possible responses. However, this is intractable, and various search algorithms adopt different sampling strategies to approximate the full search while staying close to the original model. BoN samples $N$ sequences from the LLM and thereby implicitly maintains the constraint. Monte Carlo Tree Search \cite{liu2023don} still uses the initial LLM to sample sequences, but uses a Value function to direct the search into high-reward responses. These algorithms and their variants have been shown to achieve remarkable performance but are expensive to compute \cite{nakano2021webgpt, eisenstein2023helping}. 

Another notable line of work \cite{dathathri2019plug, krause2020gedi} introduces a family of search algorithms that augment the token sampling with a classifier in order to find high-reward sequences. In each decoding step, they augment the LLM distribution with the classifier predictions and sample from it. The closest to our work among these algorithms is FUDGE \cite{yang2021fudge}, which trains the classifier on partial responses from an offline dataset. In contrast to these prior methods, our work leverages the RL framework to train the augmented Value estimator so that it leads to the optimal solution to the constrained optimization problem.

\paragraph{Reinforcement Learning for Language Models.}
The most commonly used RL algorithm for tuning LLMs is Proximal Policy Optimization (PPO, \cite{schulman2017proximal}) and its variants  \cite{ramamurthy2022reinforcement, wu2023pairwise}. Most of these are actor-critic algorithms \cite{sutton2018reinforcement}, which iterate between learning a Value estimator for the current policy and using this estimator to improve the policy. This bi-level process can lead to a suboptimal policy, as has been observed empirically \cite{gao2023scaling, mudgal2023controlled}. Unlike these methods, our algorithm utilizes a closed-form solution to extract the optimal constrained policy given a Value function, bypassing the policy optimization step. In addition, we learn the Value function of a fixed initial policy, thereby avoiding all of the aforementioned difficulties in optimization. \label{sec:ppo_problems} Of most similarity to our work is ILQL \cite{snell2022offline}, which proposes a Q-learning algorithm for steering LLMs. However, their work focuses on the offline RL setting, while ours assumes online interactions with the environment (i.e., prompts and reward function). We further discuss these similarities and differences in Appendix \ref{sec:appx_B}. 

Concurrently and independently of our work, Controlled Decoding \cite{mudgal2023controlled} explores a similar method to use the Q function to guide the decoding process. 

\section{Method}
\label{methods}

\subsection{Problem Definition.} Consider the following finite-time Markov Decision Process (MDP) for a generative model $\pi$ (``policy"): at time step $t$, the policy observes a state $s_t$, defined as the text prompt $y$ concatenated with the response generated so far $x_{\leq t}$, and outputs an action $x_{t+1} \in \mathbb{V}$, where $\mathbb{V}$ is the vocabulary of tokens, $x$. Once the action is chosen, the model deterministically transitions into the next state $s_{t+1}=s_t\oplus x_{t+1}$, where $\oplus$ is the concatenation operation. The episode ends either after the policy generates T new tokens or an [EOS] token. At the end of the episode, a scalar reward is collected from the reward model $r(s_T): \mathbb{V}^T \to \mathbb R$ where $s_T$ is the generated response. In the following Kullback-Leibler (KL)-regularized RL setting \cite{ng2000algorithms,boularias2011relative}, we aim to learn a new policy that maximizes the expected reward while keeping the distance from the initial $\pi_0$ under a threshold $\epsilon$:

\begin{equation}
\label{eq:rl_problem}
\max_\pi {\mathbb E_{s_T\sim\pi} [r(s_T)]} \quad s.t \quad D_{KL}(\pi||\pi_0)<\epsilon
\end{equation}
where $D_{KL}$ is the KL divergence. 

This work assumes access to the output distribution of $\pi_0$, but not necessarily to its weights. This assumption holds true in the case of open-source models and proprietary models with access to logits through API.

\subsection{From Reward Optimization to Guided Decoding.}
Prior works tackled the optimization problem in Equation \ref{eq:rl_problem} by directly optimizing a language model to be the solution $\pi$. As aforementioned, solving this constrained optimization problem under the actor-critic formulation can be complex and unstable. 
As an alternative, we look into the following closed-form solution to the KL-regularized RL problem \cite{peters2007reinforcement, peng2019advantage}.

\begin{proposition} \label{pro:bayes_formulation} Let the expected reward of being in state $s_t$, and choosing token $x_{t+1}$ be the Q-value $Q^{\pi_0}(x_{t+1}|s_t)=\mathbb E_{\mathsf{s_T}\sim\pi_0}[r(\mathsf{s_T})|\mathsf{s}_{t}=s_{t}, \mathsf{x}_{t+1}=x_{t+1}]$. The solution to the optimization problem in Equation \ref{eq:rl_problem} is:
\begin{equation}
\label{eq:solution}
    \pi(x_{t+1}|s_t) \propto \pi_0(x_{t+1}|s_t) e^{\beta Q^{\pi_0}(x_{t+1}|s_t)}
\end{equation}
$\beta$ is a scalar dependent on $\epsilon$ and independent of $x_{t+1}$. 
\end{proposition}

 This proposition implies that instead of learning $\pi$, we can estimate the optimal solution $\pi$ by learning $Q^{\pi_0}$. Oftentimes, learning $Q^{\pi_0}$ for a fixed policy $\pi_0$ is simpler and stable. Therefore, it allows us to solve the KL-regularized RL problem more easily without having an explicit model for $\pi$.

\begin{figure*}[h]
    \centering
    \includegraphics[width=\linewidth]{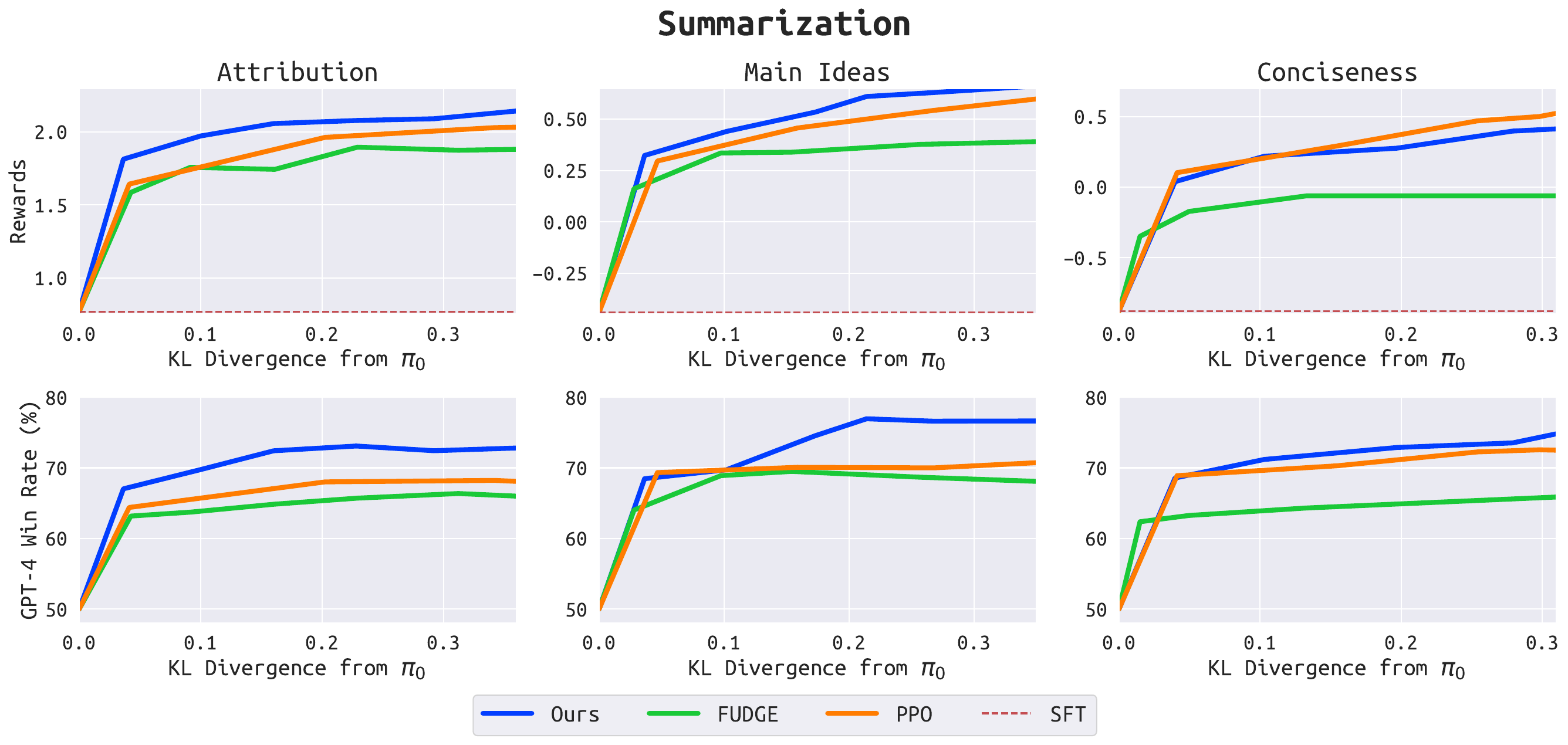}
    \vspace*{-7mm}
    \caption{The KL-performance trade-off for both reward and win rate over SFT, for different summarization objectives. \ourmethod outperforms the baselines in both Attribution and Main Ideas. In Conciseness, PPO achieves a higher reward but a significantly lower win rate. These results show that \ourmethod maximizes the reward better without reward hacking as much as PPO.}
    \label{fig:seahorse_reward_winrate}
    \vspace*{-3mm}
\end{figure*}

Learning a Q-value estimator for a given policy is a well-studied problem \cite{sutton2018reinforcement}. However, in language modeling, there are unique opportunities and difficulties. Unlike most RL environments, where the action space is either continuous or low-dimensional discrete \cite{zhu2021overview}, the action space of language models is the set of all possible next tokens $\mathbb{V}$, which is usually in the tens of thousands. Existing algorithms for Q-value estimation have been shown to be ineffective in such large discrete action spaces \cite{dulac2015deep}, thus presenting a significant challenge in applying RL algorithms to natural language tasks \cite{he2015deep, guo2021efficient, ramamurthy2022reinforcement}. To overcome this challenge, we take advantage of another unique property of the language generation task: \textit{deterministic transitions}. Given the current state and the next action, the next state is simply the concatenation of the generated tokens. This property allows us to draw the equivalence between the Q-value of a state-action pair and the value function:
\begin{equation}
\label{eq:q_and_value}
    Q^{\pi_0}(x_{t+1}| s_t)=V^{\pi_0}(x_{t+1}\oplus s_t)
\end{equation}
This allows us to re-write Equation \ref{eq:solution} as,
\begin{equation}
\label{eq:solution_v}
    \pi(x_{t+1}| s_t) \propto \pi_0(x_{t+1}| s_t) e^{\beta V^{\pi_0}(x_{t+1}\oplus s_t)}
\end{equation}
As seen in Eq. \ref{eq:q_and_value}, the value function does not depend on the action space size. Therefore, when the action space is large, it can be learned more easily than a Q-value function. The trade-off of this lower learning complexity is the inference cost of predicting $V^{\pi_0}(x_{t+1}\oplus s_t)$ for all $x_{t+1} \in \mathbb{V}$. In other words, the value estimator needs to predict the expected return of all possible subsequent states. We discuss this trade-off in more detail in Appendix \ref{sec:appx_B}.

\subsection{Value Augmented Sampling}
\label{ssec:VAS}
In the following section, we present Value Augmented Sampling (\ourmethod). Equation \ref{eq:solution_v} presents an elegant but computationally expensive solution to the KL-regularized RL problem. 
To realize it into a practical algorithm, we propose the following approximation. At every decoding step, we first obtain $k$ tokens with the highest output probabilities under our base policy $\pi_0$. Then, we calculate the value estimation for only these $k$ tokens and sample the subsequent action according to the following distribution:

\begin{equation}
\label{eq:approximation}
    \pi(x_{t+1}| s_t) \propto \begin{cases}
\pi_0(x_{t+1}| s_t) e^{\beta V^{\pi_0}(x_{t+1}\oplus s_t)}&x_t\in\text{top-k}\\
\pi_0(x_{t+1}| s_t) e^{\beta \bar{V}(s_t)}&x_t\notin\text{top-k}
\end{cases}
\end{equation}

With $\bar{V}(s_t)=\frac{1}{k}\sum_{x_t\in\text{top-k}}{V^{\pi_0}(x_{t+1}\oplus s_t)}$. We have found that even when $k << \abs{\mathbb{V}}$, our proposed approximation is effective while reducing computational complexity significantly. 

Next, we will analyze the inference-time compute required by \ourmethod. Let $m$ be the compute (in FLOPS) required by an LLM to process a single token, and $n$ the compute required by a secondary model (either value estimator or reward function) to do the same. Policy optimization methods directly optimize the base policy and, therefore, only require $o(T^2\cdot m)$ FLOPS to generate a response of length $T$. BoN requires generating $N$ responses and evaluating all of them with the reward model, requiring a total of $o(N\cdot T^2(n+ m))$ FLOPS. \ourmethod requires one pass over the base policy and $k$ token evaluation with our value estimator, requiring a total of $o(T^2(m + kn))$. When $n<<m$, our method can be significantly more efficient than BoN and approach the computational cost of using a single policy.

To train our Value function, we first collect a dataset $D= \{s^i,y^i, r^i\}_i^N $ by sampling $s^i \sim \pi_0(s_T|y^i)$ and evaluating them with the reward function $r^i=r(s^i)$. Then, we use TD($\lambda$) algorithm \cite{sutton2018reinforcement, schulman2015high} to train our value estimator. Our value estimator is parametrized as a separate neural network, decoupled from $\pi_0$. This parametrization removes both the need to have access to the initial policy's weights and update them.
We provide an illustrative overview of our algorithm in Figure \ref{fig:algo}.

\section{Experiments}
We compare \ourmethod against prior works on adapting LLMs and highlight the unique advantages of our method. We first compare \ourmethod against baselines on two common text-summarization benchmarks (Sec. \ref{sec:summarization}) and multi-turn chat dialogue (Sec. \ref{sec:hh}). We additionally demonstrate \ourmethod ability to simultaneously maximize multiple reward functions \ref{sec:multi_axis}. Then, we showcase our algorithm's ability to teach black-box models (GPT-3.5) new capabilities of tool-use (Sec. \ref{sec:gpt}). Finally, we empirically demonstrate the ability to perform fine-grained control over the specified task behavior during inference in the task of LLM personalization (Sec. \ref{sec:personalization}). Finally, Sec. \ref{sec:small_model}, \ref{sec:generalization} and \ref{sec:estimator_performance} present ablation studies.

\paragraph{Experimental Setup}
Ours and baseline methods are used to align a base LLM using a reward function after supervised fine-tuning of a pre-trained model on the respective datasets.
This model will serve as the initial LLM, $\pi_0$, and will be called the supervised fine-tuned (SFT) policy. For all of the models up until Section \ref{sec:personalization}, we use LLaMA 7B (either version 1 or 2, see Appendix \ref{sec:appx_C}).
For the reward models, we use popular pre-trained models, and when not available, we train custom reward models with FLAN-T5-L (780M parameters) \cite{wei2021finetuned}. To maximize the performance of each method, we conducted independent hyperparameter searches for each method and reported the result for the best hyperparameter setting. See Appendix \ref{sec:appx_C} for more details. 

\paragraph{Evaluation}
Following \cite{gao2023scaling, rafailov2023direct, wu2023pairwise}, we conduct an evaluation using two different metrics. First, we analyze the trade-off between reward and KL divergence from the SFT model to study each algorithm's ability to maximize the specified reward. Further, due to the known `reward hacking' problem in many algorithms, we use GPT-4 as a proxy for human evaluators to conduct comparisons between models. Many works have shown that GPT-4 has a high correlation with human judgment \cite{dubois2023alpacafarm, alpaca_eval}. We report win rates against responses generated by the SFT model and conduct direct comparisons with our baselines.

\begin{figure*}[h]
    \includegraphics[width=\linewidth]{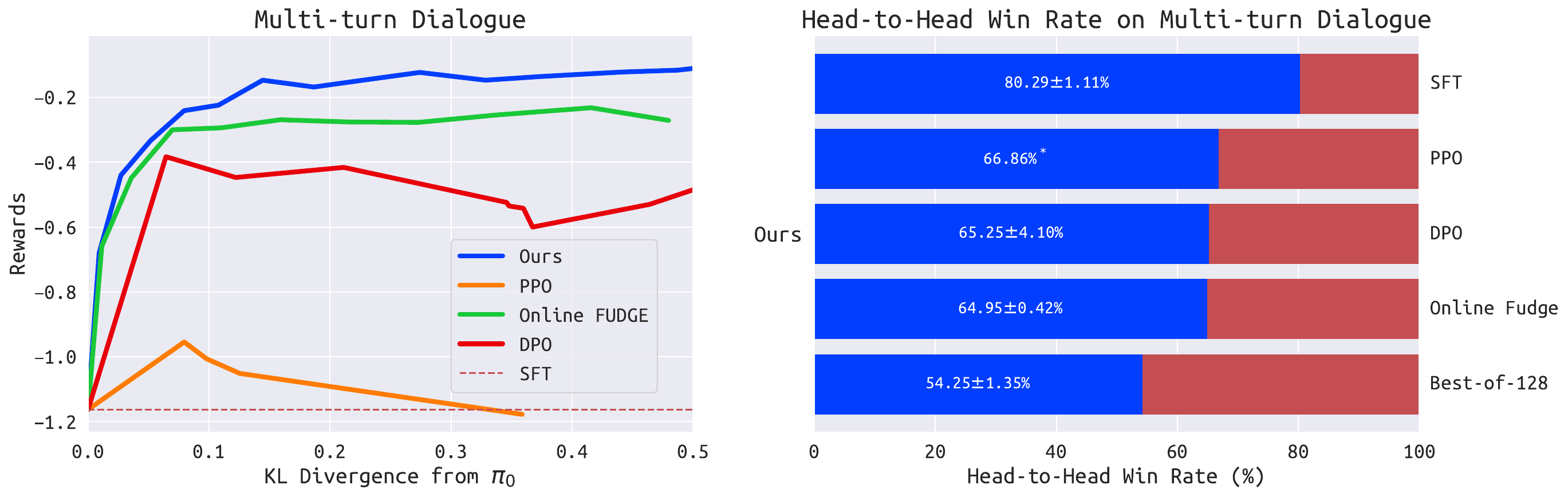}
    \vspace*{-5mm}
    \caption{Performance on the Anthropic's HH dataset using LLaMA-2 7B model. \ourmethod outperforms strong baselines like DPO and achieves comparable results to Best-of-128. We report the average win rate and standard deviation (error bar) across three random seeds for each of the methods.}
    \label{fig:hh_rewards_winrate}
    \vspace*{-3mm}
\end{figure*}

\subsection{How does VAS perform on RLHF objectives?}
\subsubsection{Summarization}
\label{sec:summarization}

\paragraph{Task}
We begin with the task of improving summary generation, one of the standard benchmarks well-studied in literature \cite{ouyang2022training, wu2021recursively, ramé2023rewarded}. For this task, we use the SEAHORSE dataset \cite{clark2023seahorse}. This dataset contains 96K pairs of texts, summarizations, and human labels along multiple dimensions of quality. We focus on three measures: (1) \textbf{Attribution}: whether the information in the summary is fully attributable to the source article, (2) \textbf{Main ideas}: whether the summary captures the main idea(s), and (3) \textbf{Conciseness}: whether the summary concisely represents the source article.

\paragraph{Baselines}
We compare \ourmethod against PPO and FUDGE on the KL vs. reward and win rate. In the GPT-4 evaluations, we additionally include MCTS search on top of the PPO model, as suggested in \citep{liu2023don} and Best-of-128. For MCTS, we chose hyperparameters that will result in the same amount of FLOPS for inference ($S=11$, $K=20$) as VAS with $K=20$.

\begin{table}[t]
\centering
\begin{tabular}{@{}lllll@{}}
\toprule
Model   & Attribution      & Main       & Concise-      & Avg. \\
   &       & Ideas      &  ness     &  \\\midrule
SFT     & 73.1\% & 77.0\% & 74.9\% & 75.0\% \\
PPO     & 64.8\% & 59.3\% & 56.2\% & 60.1\% \\
PPO + MCTS  & 62.2\% & 54.7\% & 54.0\% & 57.0\% \\ 
FUDGE   & 59.2\% & 61.1\% & 58.8\% & 59.7\% \\
BoN-128 & 49.7\% & 48.3\% & 49.2\% & 49.1\% \\ \bottomrule
\end{tabular}
\caption{Head-to-head comparison between our algorithm to various baselines, as judged by GPT-4. Our model outperforms PPO and FUDGE and is almost on par with Best-of-128, while having lower inference compute cost.}
\label{tab:seahorse_head2head}
\end{table}

\paragraph{Results}
We first analyze the results in Figure \ref{fig:seahorse_reward_winrate}, which compares the reward and win rates against SFT results. As illustrated in the left two columns, we outperform PPO and FUDGE on Attribution and Main Ideas. We maximize the rewards given any KL divergence constraint better and summarize better overall (as determined by GPT-4). It is important to mention that FUDGE does not try to maximize the reward directly, but instead trains a classifier directly on the labeled data, which can explain its considerably low reward scores. However, the GPT-4 win rate shows that \ourmethod learns an overall better summarization ability. This highlights the benefits of using a value estimator over a classifier to augment the token-level distribution. Unlike FUDGE, PPO shares the same learning objective as \ourmethod, but achieves worse results both in reward and win rate. We attribute the reason for the performance gap to the PPO training instabilities mentioned above (Sec. \ref{sec:ppo_problems}), which often leads it to converge to a sub-optimal solution. On the last axis of Conciseness, although PPO achieved higher rewards than \ourmethod, its generated summaries are judged to be worse by GPT-4. This is a tangible example of 'reward hacking', a common problem with RL-based methods that directly optimize the policy.
We further evaluate the model's summarization ability by directly comparing the generated responses from different models against each other in \cref{tab:seahorse_head2head}. The results show that \ourmethod clearly outperforms the baselines, with an average 60\% success rate. Performing MCTS on top of PPO improves the results a bit, but \ourmethod still outperforms it. More importantly, we approximately match the performance of Best-of-128, a very strong but computationally expensive method. \looseness=-1

\subsubsection{Chat Dialogue}
\label{sec:hh}
\paragraph{Task}

The second task is to improve the LLM's ability to engage in multi-turn dialogue as a chat assistant. To this end, we use Anthropic's Helpfulness and Harmlessness (HH) dataset \cite{bai2022training}, which contains $\sim$161K examples of conversations between a human and an assistant. Each conversation is paired with two responses, one of which is labeled as preferred. The preference is based on which is more informative, honest, and safe. \looseness=-1

\paragraph{Baselines} On top of the previous baselines of PPO and FUDGE, we add another strong baseline, Direct Policy Optimization (DPO) \cite{rafailov2023direct}. DPO is an offline contextual bandit algorithm that learns a policy directly from preference data. Although being a strong baseline, DPO is designed specifically for a preference dataset with pairs of responses and, as such, is unsuitable for the other tasks considered in this work. Contrary, FUDGE is not designed for pair-wise preference learning. Therefore, we create a variation of FUDGE (``Online FUDGE'') as a baseline for this task. Instead of learning a prefix classifier from offline data, Online FUDGE learns a Value estimator from online data and reward function by regarding it as a regression task. Therefore, it can also be seen as a TD(1) version of \ourmethod. For GPT-4 evaluations, we additionally evaluate against Best-of-128. \looseness=-1

\paragraph{Results}
As in the summarization task, we study the KL-reward frontier and evaluate the quality of the generated responses as a chat assistant with GPT-4. As illustrated in the left plot of Figure \ref{fig:hh_rewards_winrate}, \ourmethod attains the best KL-reward trade-off among all the models. DPO's weak performance on this metric originates from the fact that it does not learn to maximize the reward, but instead learns directly from the data. On the right plot, we show head-to-head win rate comparisons against different models. Noticeably, we outperform DPO, a strong baseline tailored specifically for preference optimization, with a 55\% win rate. In addition, we show that the responses generated by \ourmethod are comparable, if not slightly better, to best-of-128. Regarding the poor performance of PPO, we report the best performance we were able to achieve. Although we experimented with three well-regarded codebase implementations \cite{alpaca, vonwerra2022trl, havrilla-etal-2023-trlx} and over 20 hyperparameter search, we found PPO to be unstable on this dataset. Prior work also reports similar difficulties in training PPO on this dataset \cite{rafailov2023direct}. This once again alludes to the algorithm’s optimization instability.

\subsection{Composing multiple alignment objectives}
\label{sec:multi_axis}
\paragraph{Task} In the real world, alignment to multiple different axes is required to model the complex set of human preferences. Thus, we further extend the evaluation of our algorithm's capabilities in solving multi-axis alignment. Specifically, we focus on the task of improving the three axes of Attribution, Main Ideas, and Conciseness simultaneously. Under the RL framework, multi-axis optimization is usually framed as maximizing a linear combination of different reward functions \cite{hayes2022practical}. To transform \ourmethod into a multi-reward optimization algorithm, we can take advantage of the fact that the Value is a linear function of the reward \cite{sutton2018reinforcement}. Therefore, maximizing a linear combination of rewards is equivalent to performing \ourmethod with a linear combination of the respective Value estimators. It is important to mention that these estimators do not need to be jointly trained and are combined only during inference. \looseness=-1

\paragraph{Baselines}
For baselines, we compare against Multi-Objective PPO, a PPO training where the reward is the sum of the three separate reward functions. Moreover, we compare against Rewarded Soup \cite{rame2023rewarded}, an alternative approach to extend PPO for multi-alignment. Rewarded Soup trains a PPO model for each reward separately and averages the weights of these models during inference. We conducted the experiments under the setting that the three attributes should be weighted equally.

\begin{figure}[h]
\centerline{\includegraphics[width=\columnwidth]{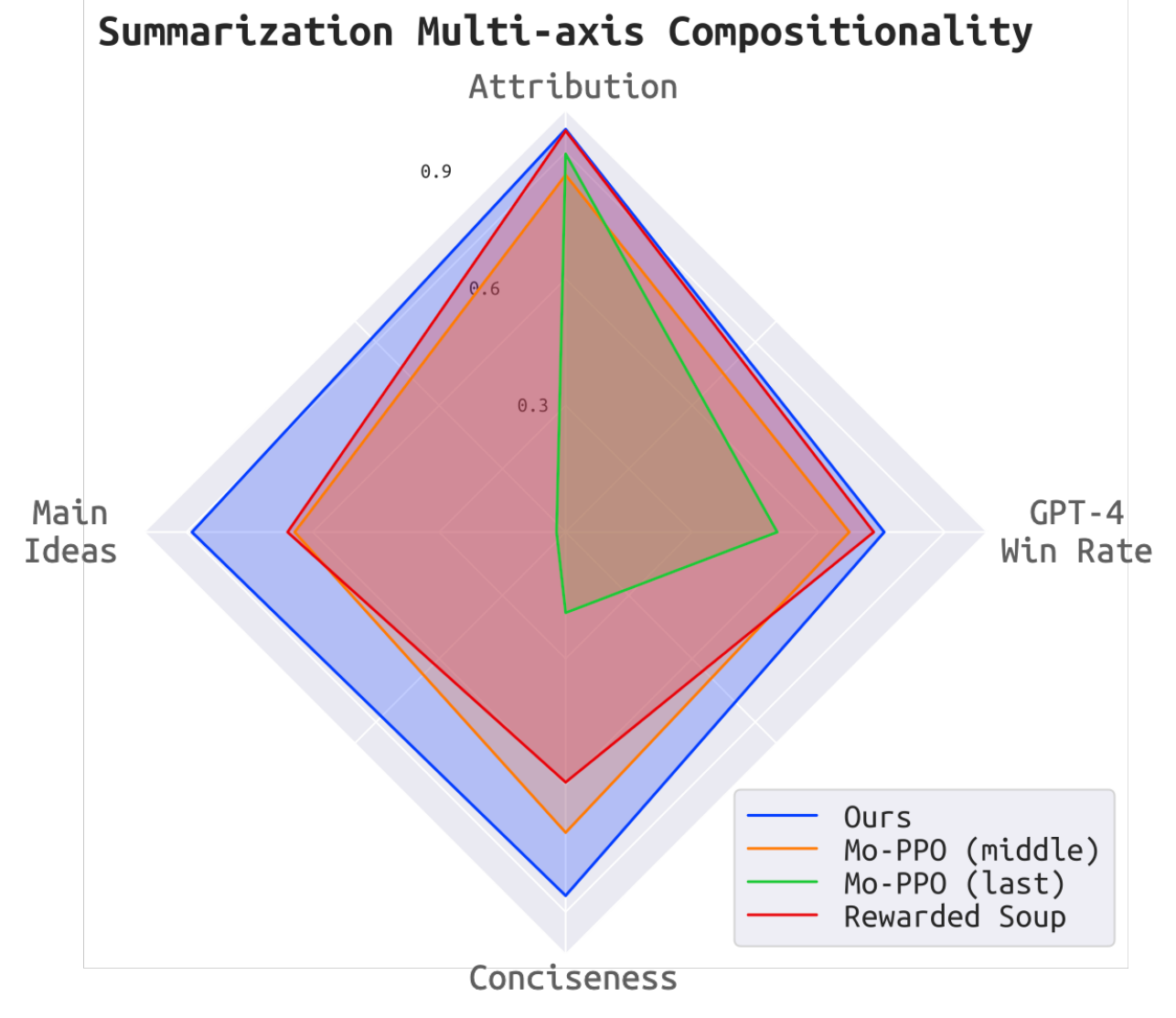}}
\vspace*{-3mm}
\caption{\ourmethod is able to maximize multiple rewards simultaneously without sacrificing performance in any axis. Mode-switching behavior can be observed during the Mo-PPO training (see the comparison between middle and last), which once again alludes to its inherent instability in optimization.}
\label{fig:seahorse_multi_rewards}
\vspace*{-3mm}
\end{figure}

\paragraph{Results}
As seen in Figure \ref{fig:seahorse_multi_rewards}, \ourmethod effectively maximizes the rewards on each axis without sacrificing the performance in another and outperforms both MO-PPO and Rewarded Soup. We observe an interesting phenomenon with MO-PPO, where the different reward models compete with each other over the course of the training, and the policy exhibits mode-switching behavior, alternating between optimizing for one reward model over the other. At one point, as shown with MO-PPO (middle), it is able to learn a reasonably good balance, but at the end of the training, as seen with MO-PPO (last), the model collapses on one of the axes. This gives another example of the instabilities in PPO training.

\subsection{VAS enables fine-grained control for personalization}
\label{sec:personalization}
\paragraph{Task}
Unlike in the previous two sets of experiments, for some behaviors, it is not always desirable to simply maximize the rewards but control it dynamically. This is important in the problem of personalization, where users have different preferences for the style of responses. Prompt engineering is a solution, but it can be time-consuming and computationally expensive to optimize the prompts to get the exact degree of an attribute \cite{pryzant2023automatic, deng2022rlprompt}. In this experiment, we show that \ourmethod is a complementary approach to prompting that allows easy tuning at inference time. This can be done by modulating the $\beta$ parameter to control the weight between the original LLM and the Value augmentation. In contrast, control at inference-time is not possible with PPO, since a new policy needs to be trained for each different configuration. We show this fine-grained control capability over the axes of \textbf{Formality} and \textbf{Verbosity}. 

\paragraph{Experimental Details}
For the initial policy, we fine-tune LLaMA-2-7B on Alpaca instruction following dataset \cite{alpaca} to get an SFT model that can follow human instructions. For Formality, we use a pre-trained reward model from \cite{formality} based on Roberta Base (125M parameters) trained to predict whether a sentence is formal or not. For Verbosity, we use a numerical function of the sequence length as the response. We first collect 15,000 different topics (e.g., data compression, fruit wines, etc.) and use this data in tandem with the aforementioned reward models to train our Value estimators, which are also LLaMA-2-7B models. 

\begin{figure}[h]
\centerline{\includegraphics[scale=0.4]{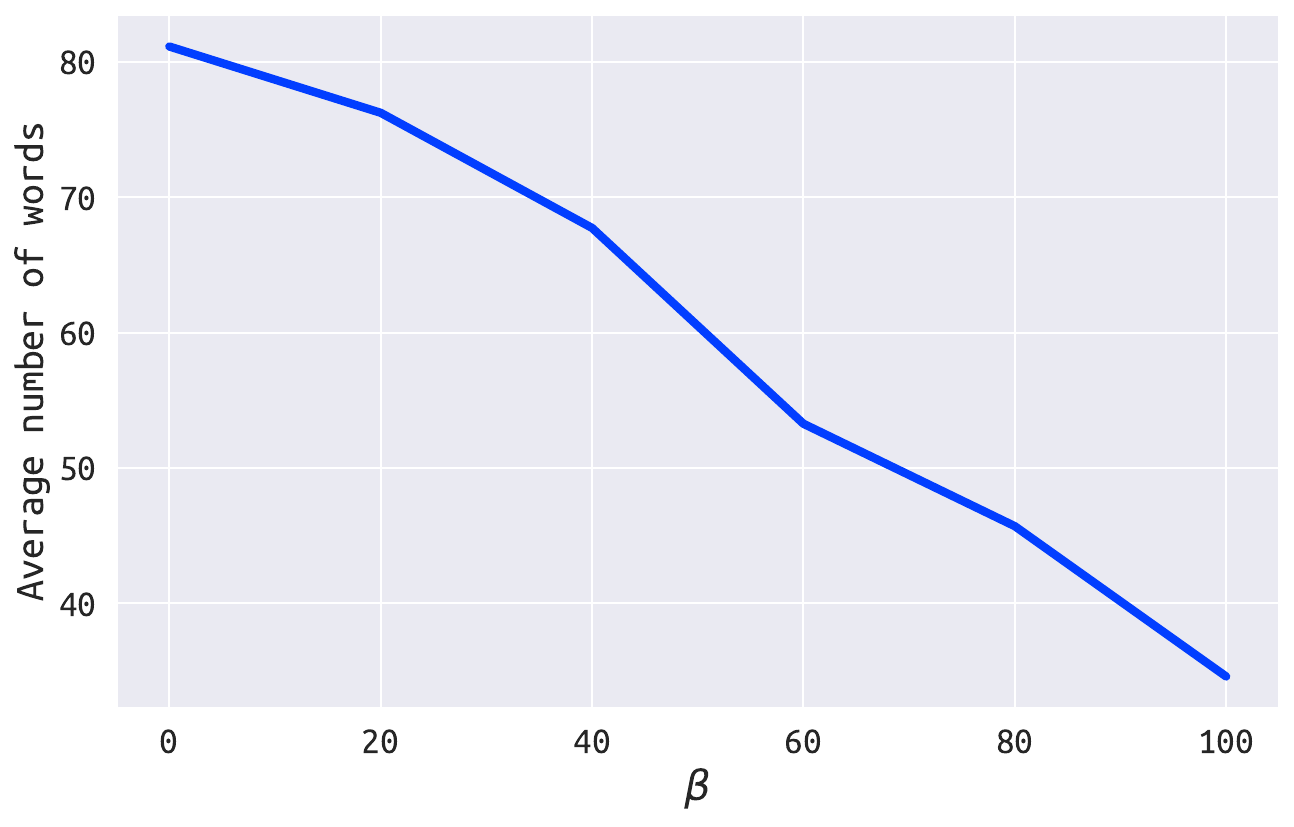}}
\vspace*{-3mm}
\caption{As we increase $\beta$ during decoding, the average number of words per response decreases monotonically. This illustrates the ease of tuning LLMs' response lengths and versatility in reaching any point in the curve.}
\label{fig:verbosity}
\vspace*{-3mm}
\end{figure}

\begin{figure*}[h!]
    \centering
    \includegraphics[scale=0.4]{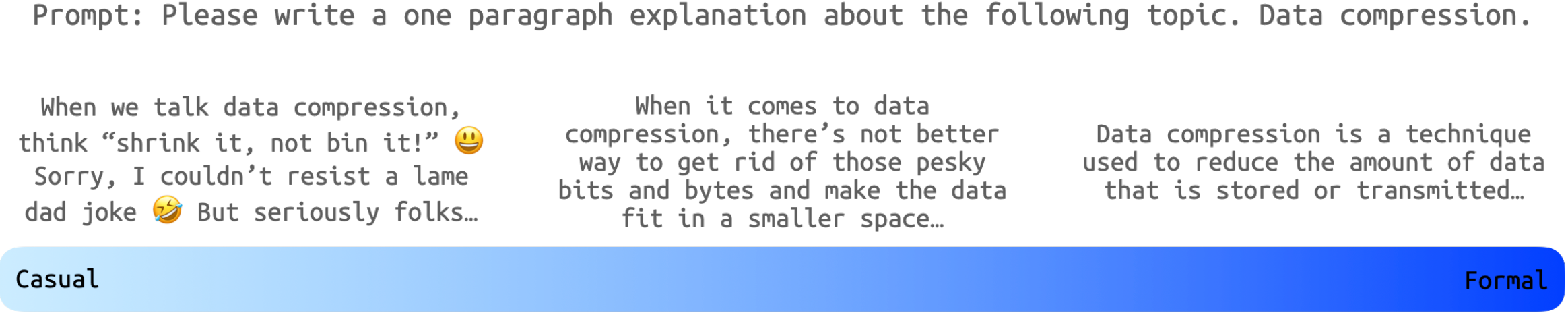}
    \vspace*{-3mm}
    \caption{Qualitative results on Formality. By simply controlling $\beta$, our model easily varies from generating casual responses with emojis on the left to generating formal, technical answers on the right.}
    \label{fig:formality_qualitative}
    \vspace*{-3mm}
\end{figure*}

\paragraph{Results}
In Figure \ref{fig:verbosity}, we vary the $\beta$, the weight of our Value estimator, when augmenting the initial model. We clearly observe that there is a monotonic decrease in the average number of words per response as we increase $\beta$. This result illustrates that not only can we intuitively and easily change the behavior of the LLM, but also our algorithm has the versatility to reach any desired point in this alignment axis. For qualitative evaluation, we showcase how the response to the same prompt evolves when varying the $\beta$ parameter for the axis of Formality. Figure \ref{fig:formality_qualitative} shows that the model starts to generate more causal responses when $\beta$ is lower and more formal, technical responses when $\beta$ is higher. Further examples can be found in Appendix \ref{sec:appx_D}.

\subsection{Teaching GPT-3.5 tool-use}
\label{sec:gpt}
An important advantage of \ourmethod is that it can work with black-box LLMs and we demonstrate this capability by teaching GPT-3.5 to use an API tool. Following \cite{xu2023tool}, we experiment with the task of learning to use \textit{Home Search} API. In this task, the model is provided an explanation of the APIs (Zero-shot setting) and asked to perform a sequence of API calls to complete a search given the user query. In the one-shot setting, we append a demonstrative example of such a query and response to the prompt. \looseness=-1

To train \ourmethod, we first collect 8K query examples and GPT-3.5 responses to these queries. Then, we used the success or fail signal as the reward and trained a Pythia-1B model as a Value estimator. At inference, we query GPT-3.5 for the top-5 logits of the next token through OpenAI's API\footnote{https://platform.openai.com/docs/api-reference/chat/create\#chat-create-top\_logprobs}, augment them with our Value estimates, and sample the token with the highest combined score. \looseness=-1

As shown in Table \ref{tab:gpt-3.5}, GPT-3.5 (zero-shot) only succeeds $\sim 17$\% of the times. Its performance improves to $62.8$\% with an in-context example. However, when \ourmethod is combined with an in-context example, we show an additional improvement of $22.3\%$. This demonstration sheds light on our approach as an \textit{complementary} approach to teaching even closed-source, proprietary models how to use custom APIs. \looseness=-1

\begin{table}[h]
\centering
\begin{tabular}{@{}lcl@{}}
\toprule
Method & Success Rate \\ \midrule
Zero-shot          & 17.4\%     \\
Zero-shot + \ourmethod & 27.9\% \\
One-shot          & 62.8\%     \\
One-shot + \ourmethod          & 84.5\%     \\ \bottomrule
\end{tabular}
\caption{Teaching GPT-3.5 how to use Home Search API. Our method, complemented by adding an example to the prompt (One-shot), is able to improve performance by $\sim 22\%$.}
\label{tab:gpt-3.5}
\vspace*{-5mm}
\end{table}

\subsection{Can we use smaller models to align bigger ones?}
\label{sec:small_model}

\begin{figure}[H]
\centerline{\includegraphics[width=\columnwidth]{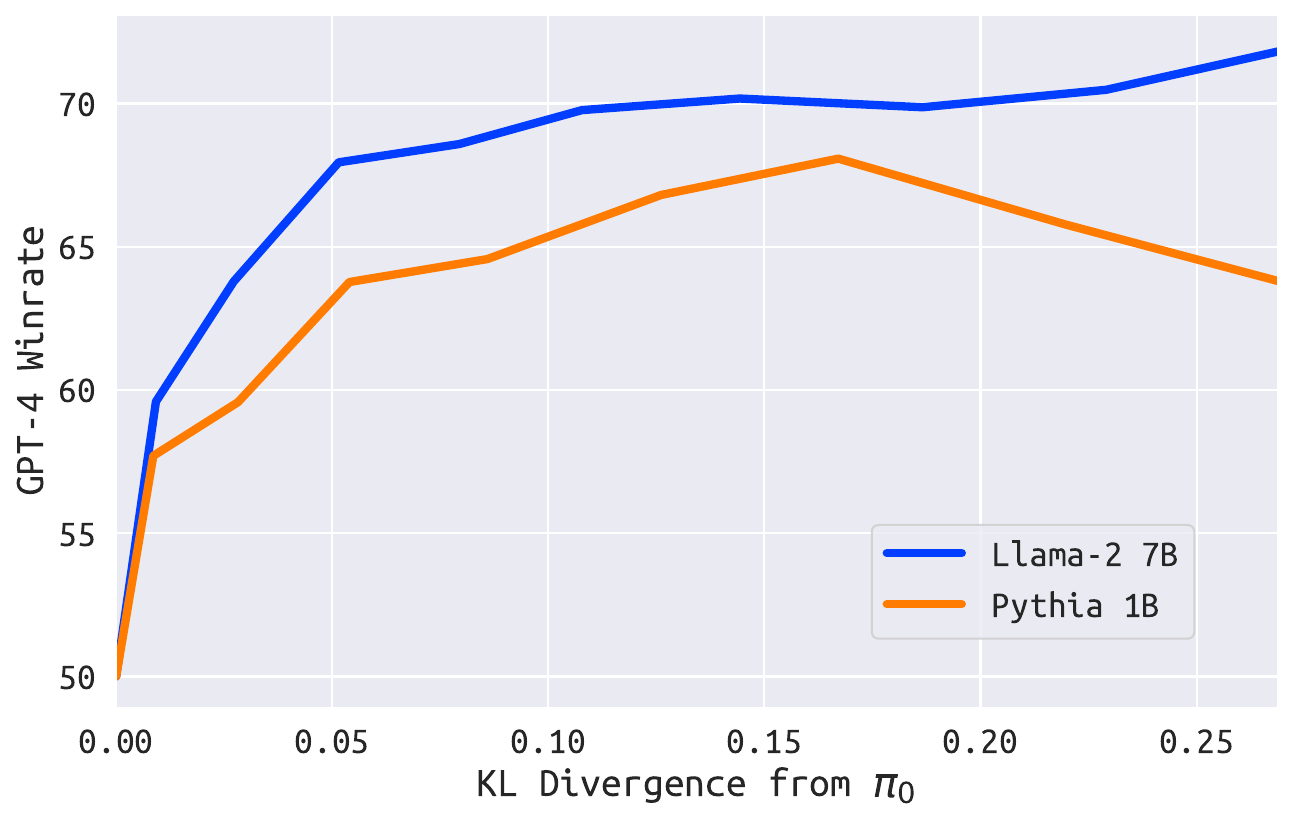}}
\vspace*{-3mm}
\caption{Comparison of different models size on the Chat Dialogue task. Smaller models achieve worse performance but require less computation during inference.}
\label{fig:model_size}
\vspace*{-4mm}
\end{figure}

\ourmethod proposes to align LLMs at inference-time by using the Value function as a search heuristic. As discussed in section \ref{ssec:VAS}, part of the potential in \ourmethod is in using smaller models. To investigate this possibility, we re-iterated the Chat Dialogue experiment, using a 1B model from the Pythia family \cite{biderman2023pythia} to guide a Llama-2 7B model. As seen in Figure \ref{fig:model_size}, the performance degrades with model size reduction, but the smaller model can still achieve a significant win rate of 68\% over SFT. We conclude that \ourmethod enables flexibility in the trade-off between compute and performance.

\subsection{Does \ourmethod generalize beyond the training data?}
\label{sec:generalization}
Because \ourmethod, specifically our Value estimator, is trained on a specific dataset of prompts and responses, a natural question is whether it generalizes beyond the training data. To test \ourmethod's generalization capability, we evaluate the performance of our \ourmethod, trained for chat dialogue on the Anthropic HH dataset, on MT-Bench \citep{zheng2023judging}. MT-Bench is a
comprehensive multi-turn benchmark that measures the ability of LLMs to engage in coherent, informative, and engaging conversations on topics of math, humanities, and more. On this benchmark, the SFT, DPO, and \ourmethod models achieve scores of 3.38, 3.91, and 4.29, respectively (higher is better). This shows that \ourmethod indeed generalizes to this new dataset. Interestingly, \ourmethod also seems to generalize better than DPO. Further, we specifically test for alignment tax on an orthogonal task of medical question answering and demonstrate that \ourmethod preserves its medical question answering abilities better than DPO. We detail the experiment and results in Appendix \ref{sec:appx_D_generalization}. 

\subsection{How does the Value estimator's accuracy affect task performance?}
\label{sec:estimator_performance}
Equation \ref{eq:solution_v} shows that leveraging our Value estimator as the search heuristic leads to the optimal solution to the KL-constrained RL problem. Since we don't have access to the true Value function and only estimate it, we question the relationship between our estimator's accuracy and the final performance. To study the Value estimator's accuracy, we generate a held-out validation set on which we generate ten different completions starting from a partial answer and compute the mean of the generations' rewards as the label. By varying the dataset size, we attain Value estimators of differing performances, and, as shown in Table \ref{tab:value_est_performance}, we observe a clear negative correlation clear negative correlation between the validation MSE and the rewards of the tuned LLM. In other words, as the Value estimator learns better estimates of the expected reward, the performance of the tuned model improves with respect to the reward model we are maximizing.

\begin{table}[h]
\centering
\begin{tabular}{@{}lll@{}}
\toprule
Dataset Size & Validation MSE $\downarrow$& Rewards $\uparrow$ \\ \midrule
32K          & 0.852          & 0.287   \\
64K          & 0.811          & 0.355   \\
96K          & 0.676          & 0.411   \\ \bottomrule
\end{tabular}
\caption{We conduct an ablation along different dataset sizes to evaluate Value estimators of varying accuracies. These results illustrate that higher accuracy (lower validation MSE) is directly correlated with performance of the tuned LLM.}
\label{tab:value_est_performance}
\vspace*{-3mm}
\end{table}

\section{Conclusion}
This work proposes an alternative approach to solving the KL-regularized RL problem to adapt LLMs via the closed-form solution. Thereby, not only does our method address the core problem of optimization instability with existing actor-critic-based RL algorithms and significantly improve the performance, but also it enables new capabilities of personalization and easy compositionality. We envision a future where people interact with models customized to their needs and preferences instead of a standardized, static one. \ourmethod serves as an important step towards this future. However, important open challenges remain along the way. An important question would be to investigate whether a Q-value function can be learned in lieu of a Value function. We explored versions of learning this Q-value estimator and encountered challenges because of the complexity of the learning problem in environments with large action spaces, as shown in Appendix \ref{sec:appx_B}. Solving this problem would significantly improve the computational efficiency and the scalability of the solution. We hope that future work continues the efforts to build a scalable, aligned, and personalized 

\section{Acknowledgments}
We thank the members of the Improbable AI lab for the helpful discussions and feedback on the paper. We also thank Phillip Isola for the initial conversations and for inspiring us in this direction. We are grateful to MIT Supercloud and the Lincoln Laboratory Supercomputing Center for providing HPC resources. The research was supported in part by NSF Graduate Fellowship, MIT-IBM Watson AI Lab, and Hyundai Motor Company. The research was sponsored by the Army Research Office and was accomplished under Grant Number W911NF-21-1-0328. The views and conclusions contained in this document are those of the authors and should not be interpreted as representing the official policies, either expressed or implied, of the Army Research Office or the U.S. Government. 

\section{Author Contributions}
\textbf{Seungwook Han} co-developed the project concept and co-led the project in all aspects of experiments and writing.

\textbf{Idan Shenfeld} co-developed the project concept and co-led the project in all aspects of experiments and writing.

\textbf{Akash Srivastava} engaged in research discussions and helped brainstorm experiments for the project.

\textbf{Yoon Kim} engaged in research discussions from its inception, helped design the experiments, and helped in paper writing.

\textbf{Pulkit Agrawal} was involved in research discussions, helped design and iterate on the experiments, conceived the positioning of the work, and overall advising.

\bibliography{main_paper}
\bibliographystyle{icml2024}

\newpage
\appendix
\onecolumn
\section{Proofs and Derivations.}
\label{sec:appx_A}
\subsection{Solution to the KL-regularized Rl problem} \label{ssec:full_derivation}
The following derivation is known as Reward-Weighted-Regression \cite{peters2007reinforcement} and is closely related to the Bayesian perspective on KL-regularized RL \cite{korbak2022rl}. As a reminder, the problem we aim to solve is:
$$
\max_\pi {\mathbb E_{s_T\sim\pi} [r(s_T)]} \quad s.t \quad D_{KL}(\pi||\pi_0)<\epsilon
$$
Where we We will start by adding a constant to the first part of the objective, transforming it into:
$$
\max_\pi {\mathbb E_{s_T\sim\pi} [r(s_T)]-\mathbb E_{s_T\sim\pi_0} [r(s_T)]} \quad s.t \quad D_{KL}(\pi||\pi_0)<\epsilon
$$
Using the Performance Difference Lemma (Lemma 6.1 from \cite{kakade2002approximately}) we can write:
$$
\mathbb E_{s_T\sim\pi} [r(s_T)]-\mathbb E_{s_T\sim\pi_0} [r(s_T)] = \mathbb E_{s_T\sim\pi}[Q^{\pi_0}(x_{t+1}|s_t)-V^{\pi_0}(s_{t})]
$$

And so our optimization problem became:
$$
\max_\pi {\mathbb E_{s_T\sim\pi}[Q^{\pi_0}(x_{t+1}|s_t)-V^{\pi_0}(s_t)]} \quad s.t \quad D_{KL}(\pi||\pi_0)<\epsilon
$$

With the additional constraint that $\pi(\cdot|s_t)$ is a probability distribution for all $t$. The constraints should hold for every state, and in particular under the state distribution induced by $\pi$. Therefore, the corresponding Lagrangian is:
$$
\mathcal{L}(\pi, \vec\lambda_1, \vec\lambda_2) = 
\mathbb E_{s_T\sim\pi} \left[Q^{\pi_0}(x_{t+1}|s_t)-V^{\pi_0}(s_t) +\lambda_1(s_t) \left( D_{KL}(\pi||\pi_0)-\epsilon\right) +\lambda_2(s_t) \left(\sum_T\pi(x_{t+1}|s_t)-1\right)\right]
$$
Where $\vec\lambda_1, \vec\lambda_2$ is state-dependent lagrange multipliers. Notice that if we consider only the decision at a specific timestep $\tau$, the expectation depends only on the policy's distribution on that specific timestep, for example the first term:
$$
\mathbb E_{s_T\sim\pi} [Q^{\pi_0}(x_{t+1}|s_t)-V^{\pi_0}(s_t)] = \sum_{x_{t+1}} \pi(x_{t+1}|s_t) [Q^{\pi_0}(x_{t+1}|s_t)-V^{\pi_0}(s_t)]
$$
Plugging it in and taking the derivative at specific timestep $t$ with respect to the probability of taking a specific action $x_{t+1}$:
$$
\frac{\partial}{\partial\pi(x_{t+1}|s_t)} \mathcal{L}(\pi,  \vec\lambda_1, \vec\lambda_2) = Q^{\pi_0}(x_{t+1}|s_t)-V^{\pi_0}(s_t) + \lambda_1(s_t) \left(\log{\frac{\pi(x_{t+1}|s_t)}{\pi_0(x_{t+1}|s_t)}}+1\right) + \lambda_2(s_t)
$$
The corresponding fixed-point solution is:
$$
\pi(x_{t+1}|s_t) = \pi_0(x_{t+1}|s_t) e^{\frac{Q^{\pi_0}(x_{t+1}|s_t)-V^{\pi_0}(s_t}{\lambda_1(s_t)}} e^{\frac{\lambda_2(s_t)+1}{\lambda_1(s_t)}}
$$
We can simplify the expression by considering all the terms that are independent of the choice of $x_t$ as part of the normalization constant, which leaves us with:
$$
\pi(x_{t+1}|s_t)  \propto \pi_0(x_{t+1}|s_t) e^{\frac{1}{\lambda_1(s_t)}Q^{\pi_0}(x_{t+1}|s_t)}
$$
\qedsymbol{}

\section{Q-Value estimation versus Value estimation.}
\label{sec:appx_B}

\begin{figure*}[h!]
    \centering
    \includegraphics[width=\linewidth]{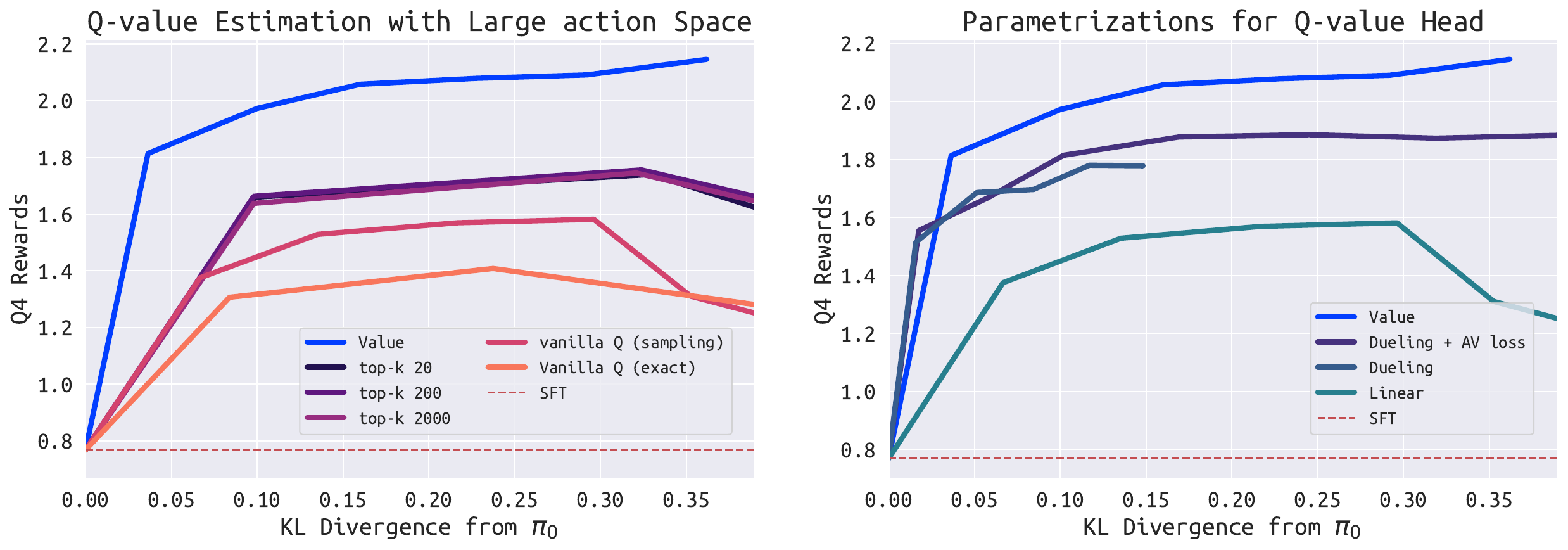}
    \caption{}
    \label{fig:q_experiments}
\end{figure*}

\subsection{Training Q Estimator for Text Generation}
In our initial formulation, as presented in Equation \ref{eq:solution}, we propose augmenting the outputs of the baseline policy with the Q-values corresponding to each token. This step is helpful for enhancing the LLM’s decoding process by considering the expected rewards associated with each potential token. Traditionally in reinforcement learning literature, Q-function estimators for discrete action spaces are parameterized as mappings from the entire state space to the action space. This typically involves a neural network architecture that processes the state as input, followed by a linear layer that outputs values spanning the action space dimension. The state space in text generation encompasses all conceivable text sequences, which is astronomically large. This is further compounded by the vast number of actions (tokens) typically involved, often numbering in the tens of thousands. Such a vast state and action space presents formidable optimization challenges. 

Given the finite size of the training dataset, rare tokens (tokens with low probability under the original LLM) are scarcely represented, resulting in minimal updates to their corresponding weights. Consequently, these rare tokens may yield arbitrary values during inference, undermining the estimator’s reliability. This is less of an issue for the state space since all input tokens share the same weights, and also, the backbone model is pretrained on a vast amount of data. To address and explore potential solutions to this challenge, we conducted a series of experiments. Through these investigations, we aim to provide insights into the complexities and practical challenges of applying Q-function estimators to text generation tasks, contributing to our decision to use a Value estimator instead. 

\subsection{Experiment Description}
All the experiments described below were done on the SEAHORSE summarization dataset, specifically with property Q4 - attribution. However, our experiments indicate that all the conclusions also hold for the other two properties in this dataset. For the Q-value estimator, we use a Llama-2 7B backbone with a linear head on top of its last hidden layer. This linear head had an output dimension of 32k, which is the vocabulary size of the Llama-2 model family. The rest of the training details are exactly the same as described in section \ref{sec:summarization}.

Our first experiment compared two ways to compute the target Q-value during Temporal Difference (TD) learning. In our experiment, we used TD($\lambda$), which constructs this value as a linear combination of several n-step bootstrap targets and the final reward $\hat{v}(s_t)=\gamma^T\lambda^T r(s_T)+\sum_{i=t}^{T-1}\gamma^i\lambda^i [\gamma V^{\pi_0}(s_{i+1})-V^{\pi_0}(s_{i})]$. Since we don't have access to the true Values of the states, we need to estimate them using the Q function. The connection between Value and Q-values that come into hand here is $V^{\pi_0}(s_i)=\mathbb E_{x_i\sim\pi}[Q^{\pi_0}(s_{i-1}, x_i)]$. Having access to the output distribution of $\pi_0$, we can calculate this expression exactly or regard the tokens in the data as samples from this expectation. Figure X (left) shows the results of these two options as '\textbf{Vanilla Q (exact)}' and '\textbf{Vanilla Q (sampling)}'. The sampling method performs much better, and we hypothesize that this is because the sampled tokens are usually those with high probability. That way it disregards the rare tokens, which usually have bad Q estimators, in the bootstrapping process.

To investigate the effect of rare tokens on the performance further, we used the Q estimator trained with the sampling-based bootstrapping and, while decoding, evaluated only $k$ tokens at each step. These \textbf{top-k} tokens were chosen as the ones with the highest probability under the base model. The results in Figure X (left) show that limiting the tokens to these with high probability helps a lot. We didn't find large performance gaps between different values of $k$  up to two thousand. After that, the performance starts to degrade. This experiment supports the conclusion that with a linear head, it is hard to learn good Q-value estimation for all tokens.

After realizing the source of the poor performance of the Q estimator, we investigated with alternative parameterization to the Q-value estimator. Since learning Q-values for rare tokens is hard, we looked for a solution that will learn a value for them that will not have an effect on the decoding. \textbf{Dueling Networks} \cite{wang2016dueling} does precisely that by decomposing the Q-value into a Value and Advantage. A dueling network has two heads, one to predict Values and one to predict Advantage, and during a forward pass it adds them together. The advantage head is initialized to zero, so rare tokens that do not change their prediction much will get an overall Q-value estimation which is just the Value. As can be seen from Figure X (right), this parameterization achieves significantly better results than the linear head. Therefore, to push the idea even further, we experimented with \textbf{Value Advantage Learning} \cite{tang2023va}, a recent learning objective that optimizes the Advantage head and the Value head separately. This represents the best results we were able to achieve using Q-value, and although they were significantly better than initial trials, they still didn't achieve the same performance as with the Value estimator.

\subsection{ILQL}
\begin{figure}[H]
\begin{center}
\centerline{\includegraphics[scale=0.4]{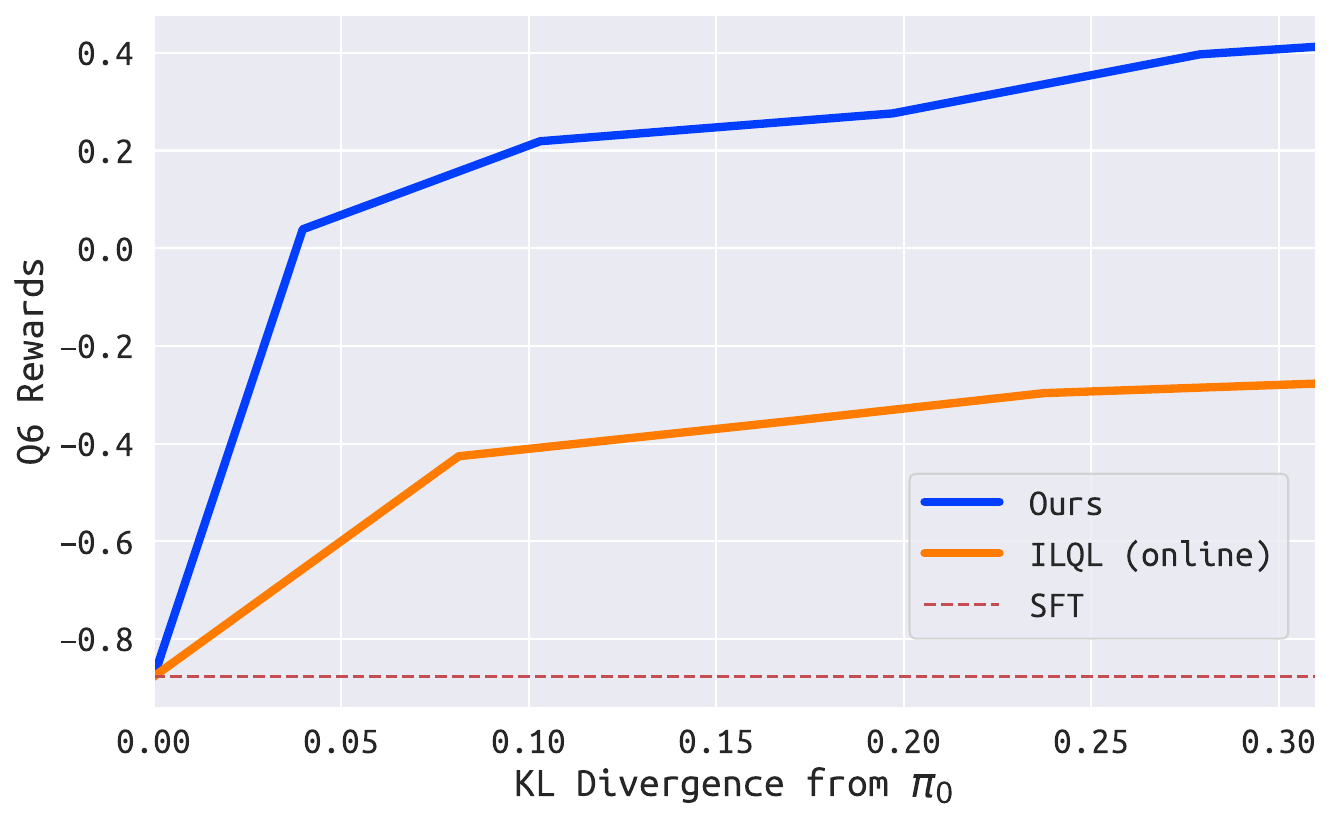}}
\caption{Comparison of \ourmethod and ILQL}
\label{fig:ilql}
\end{center}
\end{figure}

ILQL \cite{snell2022offline} is an offline RL algorithm designed for altering the behavior of LLMs given a fixed set of generated data (not necessarily from the initial LLM) annotated with a reward function. Unlike our algorithm, ILQL does not aim to learn a Q-value estimator for the current function but instead tries to learn a Q-value estimator for the optimal policy to the \textit{unconstrained } RL problem. This is the policy that solely maximizes the reward, disregarding how far its output is from those of the initial policy $\pi_0$. After learning the Q-value estimator, ILQL uses during decoding a method similar to Equation \ref{eq:solution}.
Although built for offline data, we experienced with an online version of ILQL to see if the algorithm implemented there could be useful for learning a Q-value estimator for text generation. The results in Figure \ref{fig:ilql} shows that ILQL can be used in the online setting, but is still underperformed compared to \ourmethod.

\section{Training Details and Hyperparameters.}
\label{sec:appx_C}

We attempt to ensure that comparisons to baselines are fair. In particular, we take the following steps to ensure that. First, we perform a separate hyperparameter tuning for every algorithm. For every baseline, we use either the same number of hyperparameter combinations as for our method or more. In addition, when possible, we use the codebase released together with the original paper that proposes the baseline. In cases where this baseline is not suitable for our tasks, we use a well-known baseline to ensure proper implementation. The following sections explain the details for each one of the algorithms used in this paper.

\subsection{\ourmethod Training}
In all of our experiments, we parametrize the Value estimator as an LLM backbone with a linear head. For efficient training, we use LoRA parameterization \cite{hu2021lora} and train only the low-rank adapters and linear head. The data for training was collected using the SFT model with a sampling temperature of 0.7 and then labeled by the reward function. In addition, we also collected a validation dataset by first generating a dataset of half-completed responses and, for each one, generating ten completions and averaging their reward. The training objective we use is MSE between the Value estimator output to a Value target computed with TD($\lambda$) \cite{sutton2018reinforcement}. For every task, we train three models with different learning rates and choose the one that achieves the lowest Value MSE on a validation set. For a full list of hyperparameters, see Table \ref{tab:hps_ours}. During decoding, we evaluated a range of $\beta$ values to report the KL-reward graphs. For GPT-4 evaluation, we choose the $\beta$ value that achieves the highest reward. 

\begin{table}[h]
\resizebox{\textwidth}{!}{%
\begin{tabular}{@{}ccccc@{}}
\toprule
              & Summarization                  & Chat Dialogue                                  & Tool Use       & Personalization                                                                                                   \\ \midrule
Dataset       & SEAHORSE                       & Anthropic HH                                   & Custom Dataset & Custom Dataset                                                                                                    \\
SFT Data       & All of SEAHORSE                       & Preferred Responses of Anthropic HH                                   & Alpaca Instructions & Alpaca Instructions                                                                                                    \\
Reward Model  & Trained based on T5-FLAN model & OpenAssistant/reward-model-deberta-v3-large-v2 & Success Rate   & \begin{tabular}[c]{@{}c@{}}Formality: s-nlp/roberta-base-formality-ranker\\ Conciseness: Token Count\end{tabular} \\
Value Model   & LLaMA 7B                       & LLaMA-2 7B                                     & Pythia 1B      & LLaMA-2 7B                                                                                                        \\
Lora Rank     & 60                             & 32                                             & 32             & 32                                                                                                                \\
Lora $\alpha$    & 16                             & 32                                             & 32             & 32                                                                                                                \\
Batch Size    & 32                             & 32                                             & 32             & 32                                                                                                                \\
Learning Rate & 1e-5, 5e-4, 1e-4               & 1e-4, 3e-4, 1e-3                               & 1e-3           & 1e-4                                                                                                              \\
Epochs        & 6                              & 1                                              & 1              & 8                                                                                                                 \\
$\lambda$        & 0.95                           & 0.95                                           & 0.98           & 0.95                                                                                                              \\ \bottomrule
\end{tabular}%
}
\caption{Hyperparameters for \ourmethod training.}
\label{tab:hps_ours}
\end{table}

\subsection{PPO Training}
For PPO training, we used the Alpaca Farm \cite{dubois2023alpacafarm} implementation. We parameterize both the model itself and the Value estimator with LoRA. The training data is collected through the run (as it is an on-policy algorithm) with a sampling temperature of 0.7. For the Summarization task, we have found that the KL coefficient has more effect on the performance than any other hyperparameter, and as such, did an ablation over three coefficient values. We report the results of the training (and checkpoint inside the training), which achieves the highest reward. For Multi-Objective PPO, we used the same training protocol, with the only difference being that we evaluated all three rewards during training and averaged their scores. 

We found the training for the Chat Dialogue task to be unstable, resulting in poor performance. We tried two additional implementations of PPO \footnote{https://github.com/huggingface/trl}\footnote{https://github.com/CarperAI/trlx} and tested over 20 different combinations of hyperparameters to no avail. As we mentioned before, this was reported by other research works as well \cite{rafailov2023direct}. We report the results of the best run we were able to achieve. 

For a full list of hyperparameters, see Table \ref{tab:hps_ppo}.

\begin{table}[h]
\resizebox{\textwidth}{!}{%
\begin{tabular}{@{}ccc@{}}
\toprule
                   & Summarization                  & Chat Dialogue                                  \\ \midrule
Dataset            & SEAHORSE                       & Anthropic HH                                   \\
Reward Model       & Trained based on T5-FLAN model & OpenAssistant/reward-model-deberta-v3-large-v2 \\
Policy/Value Model & LLaMA 7B                       & LLaMA-2 7B                                     \\
Lora Rank          & 60                             & 32                                             \\
Lora Alpha         & 16                             & 32                                             \\
Rollout Batch Size & 512                            & 512                                            \\
Step Batch Size    & 128                            & 128                                            \\
Learning Rate      & 1e-4                           & 5e-6                                           \\
KL coefficient     & {0.0067, 0.02, 0.05}           & 0.05                                           \\
Epochs             & 1                              & 1                                              \\
lambda             & 0.95                           & 1                                              \\
gamma              & 1                              & 1                                              \\ \bottomrule
\end{tabular}
}
\caption{Hyperparameters for PPO training.}
\label{tab:hps_ppo}
\end{table}

\subsection{DPO Training}
For DPO training, we used the codebase released by its authors. We parameterize the model with LoRA with rank 32 and $\alpha$ 32. We conducted three trainings with different learning rates $\{1e-4, 5e-5, 1e-5\}$ and reported the results of the model which achieved the highest margin on the validation set. The training was done with $\beta=0.1$ and a batch size of 64. We evaluated the model every 10K training examples for both reward and win rate against SFT. We report the DPO result as the model that achieves the best win rate against SFT.

\subsection{FUDGE Training}
For the regular version of FUDGE, used in the Summarization task, we used Transformers\footnote{https://github.com/huggingface/transformers} implementation of text classifier training and extended it for prefix classification. We parameterize the model with LoRA with rank 60 and $\alpha$ 16. We used a batch size of 32 and trained for three epochs until convergence. We performed three trainings with different learning rates $\{1e-3, 1e-4,1e-5\}$ and reported the results for the model that achieved the highest accuracy on a validation set. 
For the Online FUDGE in the Chat Dialogue task, we used the same training protocol \ourmethod and just changed the Value target to TD(1) to achieve a simple regression training. 

\subsection{Prompts for Training}
\subsubsection{SEAHORSE}
For the task of summarization, we use the following prompt for every query: 
\begin{spverbatim}
    Generate a one-sentence summary of this post.
\end{spverbatim}

\subsubsection{Anthropic Helpfuless and Harmfulessness}
For the task of multi-turn chat dialogue, we do not use any additional prompt and simply query the langauge model with the conversation between the human and assistant up until the assistant's response.

\subsubsection{GPT-3.5 Tool-use}

Following is the prompt containing brief description of the API functions that we use to query GPT-3.5:
\begin{spverbatim}
I have the following set of API:

# To set home types for search. For home buying, home_types choices are: "House", "Townhouse", "Condo", "Land", "Multi-family", "Mobile", "Co-op"; for home renting, home_types choices are: "House", "Townhouse", "Condo", "Apartment".
API.select_home_type(home_types: List[str])

# To specify whether to search homes for buying or renting. 'value' can be chosen from ['buy', 'rent']. This function must be called after setting the location and before setting any other criteria.
API.set_buy_or_rent(value: str)

# To set the minimum home size in square feet
API.set_min_square_feet(value: int)

# To set the maximum home size in square feet
API.set_max_square_feet(value: int)

# To set the floor number
API.set_floor_number(value: int)

# To set the maximum home price in dollars
API.set_max_price(value: int)

# To set the minimum home price in dollars
API.set_min_price(value: int)

# To set the location for the search area. This function must be called before setting any criteria.
API.set_location(value: string)

# Submit criterion to get search results. This function should be called after setting all the criterion.
API.search()

# To set the maximum commute time in minute to your office location, assuming the office location is already defined
API.set_max_commute_time(value: int)

# To set the number of bathroom(s)
API.set_num_baths(value: float)

# To set the number of garage(s)
API.set_num_garages(value: int)

# To set the number of balconies
API.set_num_balconies(value: int)

# To set the number of bedroom(s)
API.set_num_beds(value: int)

# To set the number of swimming pool(s)
API.set_num_swimming_pools(value: int)

-------------

Task: [USER INPUT]
 Return only the relevant API calls, not any other kind of response.
Action:
\end{spverbatim}

\subsubsection{Personalization}
For the the task of personalization, we use the following prompt for every query: 

\begin{spverbatim}
    Please write a one paragraph explanation about the following topic.
\end{spverbatim}

\section{Additional Results.}
\label{sec:appx_D}


\subsection{Learning Algorithms}

\begin{figure}[H]
\begin{center}
\centerline{\includegraphics[scale=0.4]{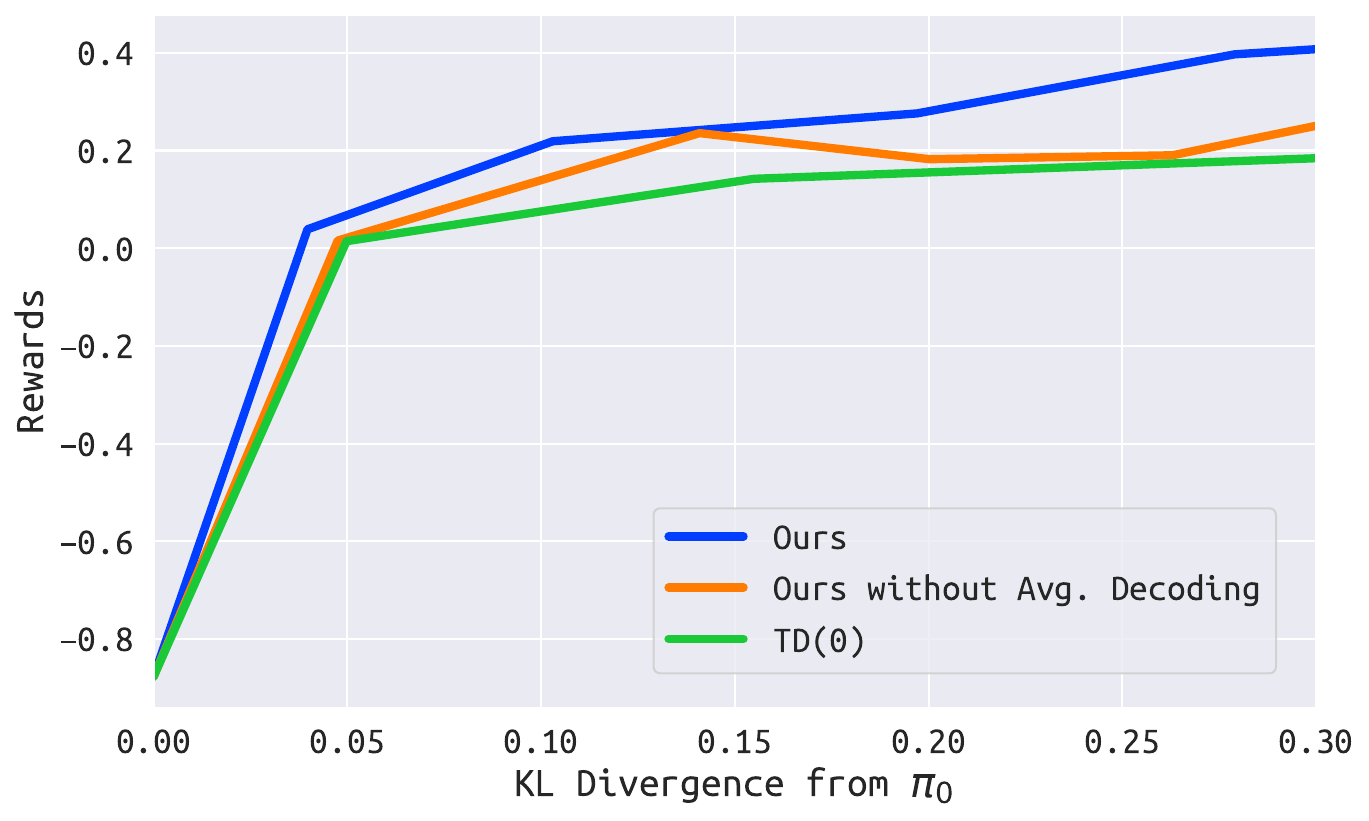}}
\caption{Abalation study on different design choices in our algorithm. The task is Summarization with Main Ideas as the reward function}
\label{fig:learning_algos}
\end{center}
\end{figure}

\ourmethod effectively leverages a Value function to guide the original LLM via token-level search, but how to best train a Value estimator and how to best guide the search (by directly augmenting the output probabilities of the LLM) are open questions. To study this question, we performed an ablation study of our method. First, we removed the augmented Value given to the tokens not in the top-k. This design choice is a heuristic to minimize the effect of our approximation by using only top-k Values. As can be seen Figure \ref{fig:learning_algos}, this causes quite a considerable degradation in performance. 

In addition, we simplified the Value function training procedure and used TD(0) instead of TD($\lambda$). This causes a small degradation in results, probably as a result of higher value estimation error.

\subsection{Varying $k$}

\begin{figure}[H]
\begin{center}
\centerline{\includegraphics[scale=0.4]{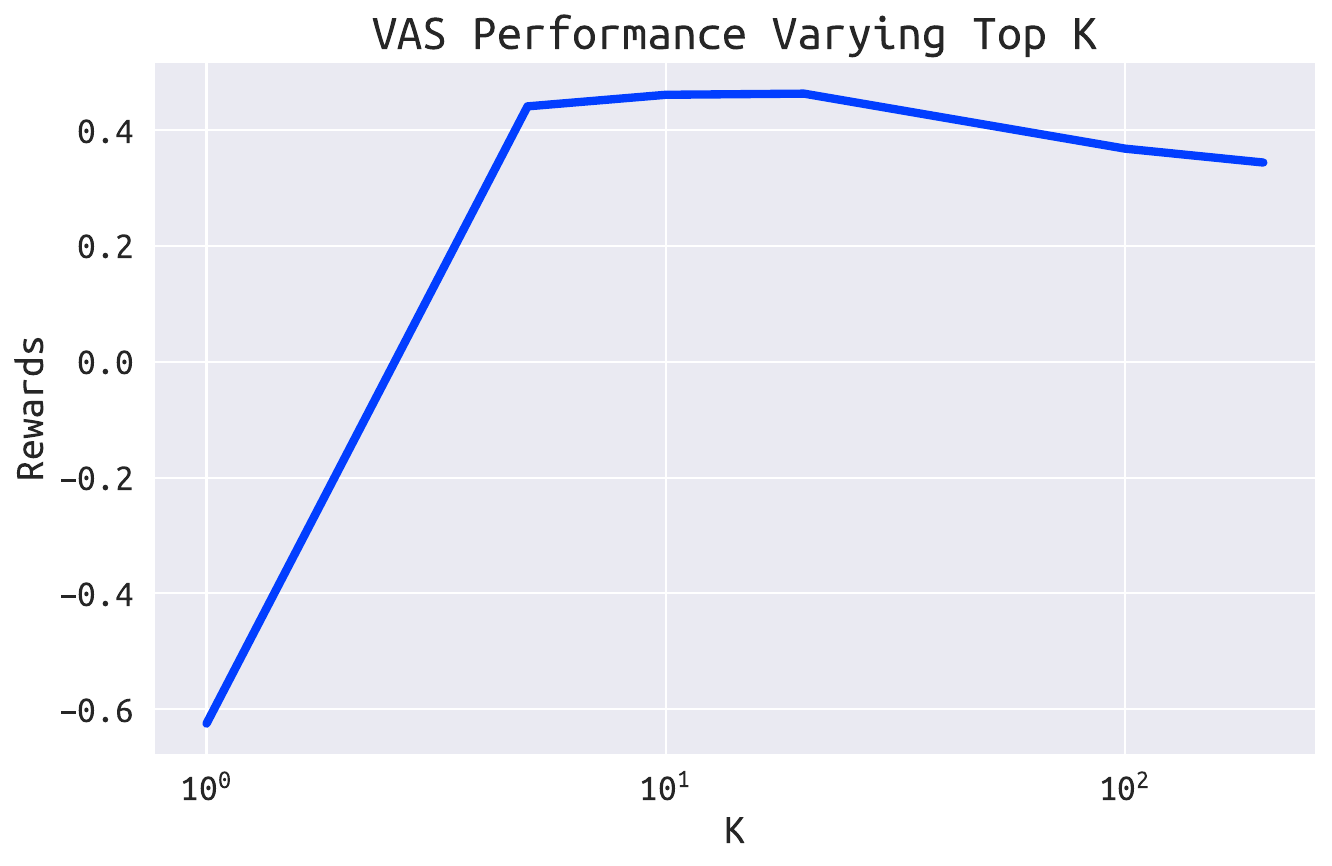}}
\caption{Varying K in \ourmethod}
\label{fig:topk_ablation}
\end{center}
\end{figure}

In Equation \ref{eq:approximation}, we propose an approximation to the closed-form solution to the KL-regularized RL problem to realize it into a computationally feasible algorithm. As seen in the equation, $k$ is a hyperparameter that directly influences the approximation and, to that end, we conduct an ablation study where we vary $k$ and evaluate the performance. We conduct this ablation on the Conciseness task for summarization on the SEAHORSE dataset and present the results in Figure \ref{fig:topk_ablation}. Interestingly, for a fixed $\beta$ of $5.0$, lower $k$ of 10 and 20 seem to be not only sufficient but more optimal than larger $k$ of 100 and 200. We conjecture that this phenomenon may be related to the fact that lower $k$ implicitly regularizes the tuned model to stay close to the initial model. However, further investigation is warranted to study how the optimal $\beta$ changes with $k$.

\subsection{Alignment Tax on Medical Question Answering}
\label{sec:appx_D_generalization}

We further assess the generalization capability or more specifically the alignment tax induced by adapting for the task of chat dialogue (Anthropic HH dataset) by testing our model on a medical multiple choice question-answering dataset. Intuitively, training on HH should not help, or in the worst case, harm the performance on this medical dataset. The SFT model achieves 33.8\% success rate. After DPO, the performance drops to 25.9\%. However, \ourmethod retains most of the original capabilities and achieves 31.1\% success rate. For the evaluation, we use \ourmethod and DPO models that achieve the same win rate on HH dataset. To evaluate the performance on MedMCQA, we used the standard protocol of looking at the tokens 1, 2, 3, 4 (multiple choice answer labels) probability.

\subsection{Inference Speed}

\begin{figure}[H]
\begin{center}
\centerline{\includegraphics[scale=0.4]{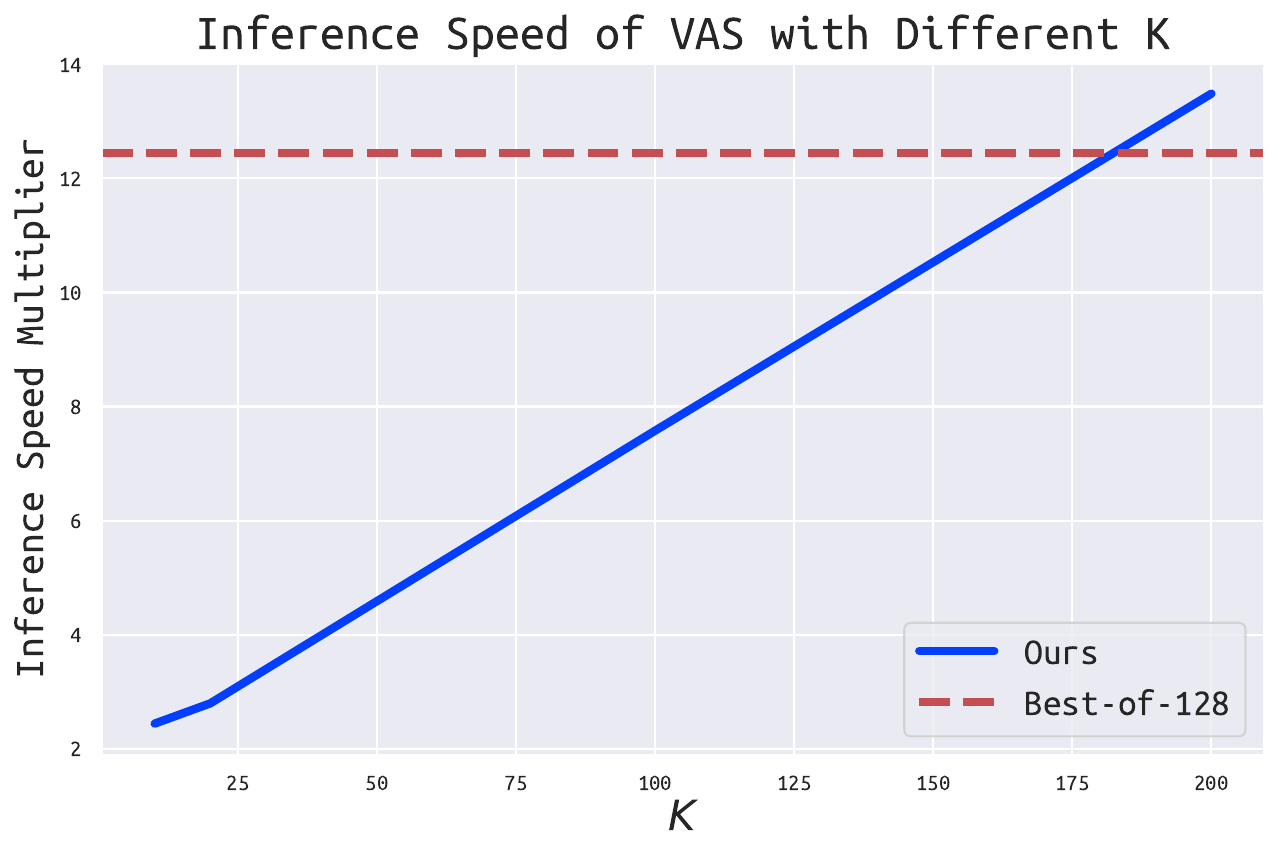}}
\caption{Varying K in \ourmethod}
\label{fig:inference_speed_k}
\end{center}
\end{figure}

We analyze the inference time with different K in our method. We conducted the analysis on a single A100 GPU. In Figure \ref{fig:inference_speed_k}, we plot the inference time multiplier (against normal decoding) for VAS with different values of $K$ and BoN with $N=128$(decoding time only, without the time needed for evaluating the reward). As expected, \ourmethod’s decoding time increases with $K$. However, the big advantage of our method is that we achieve the same performance as BoN with $N=128$ with $K=20$ ($\sim$4.4x faster than Best-of-128).

\subsection{Formality Qualitative Results}

\begin{longtable}{p{0.2\linewidth} |p{0.2\linewidth} |p{0.2\linewidth} | p{0.2\linewidth}}
\toprule
Topic                                 & Casual                                                                                                                                                                                                                                                                                                                                                                                                                                                                      & Middle                                                                                                                                                                                                                                                                                                                                                                                                                                                                                             & Formal                                                                                                                                                                                                                                                                                                                                                                                                                                                                                                                    \\* \midrule
\endfirsthead
\endhead
\bottomrule
\endfoot
\endlastfoot
Curtain                               & What's up, curtain?!? Ok, but no joke, curtains are a staple in any interior design. They not only liven up a plain-looking wall, but also provide privacy. Different fabrics and colors make a big difference. Curtains can be tailored to your own personal preferences. You can also add a lined curtain if you want a little extra warmth.                                                                                                                           & A curtain is a type of cloth that hangs from a rod or pole, usually used to cover a window or doorway. Curtains are typically made of lightweight, sheer fabrics, and come in a variety of colors and styles. They can also be used to block out sunlight, create privacy, or add a decorative element to a room.                                                                                                                                                                                  & A curtain refers to a type of window treatment used in interior design and architecture. It is made up of a fabric or other material that is hung across an opening to provide privacy, light control, and a decorative element. Curtains can be made of a variety of fabrics, such as silk, cotton, or linen, and can be used to create a range of different styles and effects.                                                                                                                                         \\ \hline
Perhipheral Vascular Disease          & What if there's no blood flowing through your arms and/or legs?                                                                                                                                                                                                                                                                                                                                                                                                             & Peripheral vascular disease (PVD) is a condition that occurs when the blood vessels outside of the heart and brain become narrowed or blocked. This can lead to reduced blood flow to the legs and feet, causing pain, numbness, and tingling. Symptoms can also include coldness, skin color changes, and ulcers.                                                                                                                                                                                 & Peripheral vascular disease is a condition that occurs when the arteries and veins in the arms, legs, or other parts of the body become narrowed due to the buildup of fatty material, causing reduced blood flow. It is most common in older adults, and can lead to pain, numbness, and discoloration of the affected areas.                                                                                                                                                                                            \\ \hline
Seasonal garden maintenance checklist & The seasonal gardening season is just getting started! There's so much to do, but don't worry - we've got you covered. \textbackslash{}n\textbackslash{}nFirst, you should check out the weather forecast. If it looks like a big storm is coming, then you should cover your seedlings and bring your potted plants indoors. \textbackslash{}n\textbackslash{}nThen, you should weed your lawn and fertilize your lawn and gardens.                                        & A seasonal garden maintenance checklist is an important part of keeping your garden healthy and looking its best. It is essential to check on the plants regularly to ensure they are getting enough water and nutrients. Pruning and trimming should be done to keep the plants healthy and looking neat. It is also important to fertilize the plants to help them grow and produce healthy flowers and vegetables. Finally, weeding should be done to keep the garden free from unwanted plants & A thorough seasonal garden maintenance checklist involves assessing the overall condition of the garden, identifying any necessary repairs or improvements, and implementing a routine of care and upkeep. It is important to inspect the soil and identify any pests or diseases, as well as check the health of the plants and trees. Additionally, it is important to regularly remove dead plants and branches, as well as clear out any debris and excess growth. It is also important to water the plants regularly \\ \hline
Mortgage interest deduction           & When it's time to get rid of the debt that's been piling up and you don't want to get stuck paying too much to lenders who are just out to make a buck, then you should definitely be looking into getting a mortgage and mortgage refinance to help you out. But wait! You can still save some bucks by deducting the interest you owe on your mortgage!                                                                                                                   & The mortgage interest deduction is a tax incentive available to homeowners in the United States. It allows homeowners to deduct the interest paid on a mortgage loan from their taxable income, reducing the amount of taxes they owe. The deduction is typically taken on Schedule A of Form 1040 and is capped at \$750,000 of debt. The deduction is beneficial to homeowners                                                                                                                   & A mortgage interest deduction is a tax benefit available to homeowners in the United States. It allows taxpayers to reduce their taxable income by the amount of mortgage interest paid on their home. The mortgage interest deduction is available for mortgage interest paid on primary and secondary homes, as well as investment properties.                                                                                                                                                                          \\ \hline
Hybrid Vehicles                       & There's nothing like getting the feel of the open road and the open skies! But, did you know that there's a new way to do just that without the extra guilt of a big carbon footprint? Hybrid vehicles are the perfect middle ground between electric and traditional combustion-engine cars. They still use fossil fuels, but they can also run on batteries. \textbackslash{}n\textbackslash{}nSo, you can still enjoy that classic driving experience you know and love. & Hybrid vehicles are a type of automobile that uses two or more sources of power to propel the vehicle. This can include a combination of an internal combustion engine and electric motor, or a battery and an internal combustion engine. Hybrid vehicles are more efficient and environmentally friendly than traditional vehicles, as they use less fuel and produce less emissions. They also offer improved performance and are easier to maintain than traditional vehicles.                 & A hybrid vehicle is a vehicle that combines the benefits of a gasoline-powered car with the benefits of an electric car. It is powered by both an electric motor and a gasoline engine, allowing it to run on both electricity and gasoline. Hybrid vehicles are much more fuel efficient than gasoline-powered cars, and they produce fewer emissions. They are also quieter and smoother than gasoline-powered cars, making them ideal for city                                                                         \\* \bottomrule
\caption{Additional qualitative results on Formality that show how our method can vary the model's responses from casual to formal with ease by changing $\beta$.}
\label{tab:formality_additional_qualitative}
\end{longtable}

\section{GPT-4 Evaluation Protocol}

As a proxy of human judgment, we follow the GPT-4 evaluation protocol from AlpacaEval\cite{alpaca_eval} and evaluate the models' generated responses. We now describe the prompts that we use for the summarization and multi-turn chat dialogue tasks.

\subsection{Summarization Prompt}
\begin{spverbatim}
    <|im_start|>system
You are helpful assistant whose goal is to simulate Diana's preferred output for a given instruction.
Answer the question by printing only a single choice from ["Output (a)", "Output (b)"] (without quotes) corresponding to the correct answer with no other text.

Here's some information about Diana:
- Diana loves literature and is highly cultivated
- Diana is a grammar freak
<|im_end|>
<|im_start|>user
## Annotation Guideline
In this task, we will ask you to select the preferred output AI model's responses to instructions.

You will read a batch of examples, which are composed of the following:

1. an Instruction we give to the AI system
2. an Input that is provided along with the instruction
3. Output (a), the first output from the AI system
4. Output (b), the first output from the AI system

Your task is to decide which response is better for each example. You should answer using only Output (a) or Output (b) depending on which response is better.

## Annotation Example
To help you understand the annotation task, we provide a single example below.
I will give an explanation for the correct answer, but you should only answer with the preferred output.

### Example 1

#### Instruction 1:
Write a design document to describe a mobile game idea.

#### Input 1:
Game Idea: A battle royale game with a superhero and villain theme.

#### Output (a) for example 1:
Design Document for a Mobile Game Idea:

1. Introduction:
A battle royale game with a superhero and villain theme.

2. Game Play:
The game is played on a large map with buildings, streets, and structures spreading across the map. Players converge on the center to compete in a winner-take-all showdown. The game includes traditional battle royale elements such as scavenging, looting, and trying to survive until the end. However, there is also a hero system in place in which players can choose to take on the role of a superhero or villain for the game. This adds some variety and variety of tactics as players use their superpowers to wreak havoc or use their villainsy to their own advantage.

3. Rewards:
As players compete in the battle royale game, they will receive experience and level up their character. These levels reward players with extra health, shields, and other all important resources. Players can choose to customize their character in various ways to make them stand out from the rest as well.

4. Levelling up:
At the end of each game, players will level up their character and receive various bonuses depending on the level. These bonuses could range from extra health to increased sprint speed or more damage while in combat. As players gain levels, they can choose to upgrade their character with more powerful abilities and powers.

5. Challenges:
Players can also complete

#### Output (b) for example 1:
The newly launched mobile game offers a thrilling battle royale experience with a superhero and villain theme. Players jump out of their cars and go head-to-head in a fight to the finish line. Each round features a different set of powerful superheroes and villainous villains, and the winner is the player who survives the longest. Along the way, they can collect health items and upgrade their powers to become the ultimate superhero or villain. The game offers a lively and vibrant art style and a visually appealing plot to keep the players engaged until the very end.

#### Which is best for example 1, Output (a) or Output (b)?:
Output (a)

Indeed, Output (a) is a design document as asked in the instruction and gives more useful details than Output (b).

## Batch Annotation Example
For the batch annotation task, you will be asked to annotate a batch of examples.
I.e. you will read all the examples first and then need to select the preferred output.
Here's an example of how you should be formatting your answers for a batch of 3 examples.
Note that I will also give an explanation for the correct answer, but you should only answer with the preferred output.

## Example 2

### Instruction 2:
Find the average age of the following people

### Input 2:
John (17), Laura (18), Robert (19), Edward (20).

### Output (a) for example 2:
The average age of the people in the given situation is 20 years old.

### Output (b) for example 2:
The average age of the people listed is 19.5 years.

## Example 3

### Instruction 3:
Generate a to-do list with 5 tasks related to the following.

### Input 3:
keeping a neat room

### Output (a) for example 3:
Make beds
Organize closet
Shoot some hole in the walls
Dust lamps and furniture
Tidy drawers and shelves

### Output (b) for example 3:
chores:
1. Clean up after yourself and put clean clothes away.
2. Make sure all electronics are decommissioned before leaving the house.
3. Remove all personal items and clutter from your room.
4. Vacuum and mop the floor.
5. Tidy up your desk and put all inboxes on hold.

## Example 4

### Instruction 4:
Rewrite the following sentence in a polite way

### Input 4:
Give it to me now.

### Output (a) for example 4:
Could you please give it to me, pending verification?

### Output (b) for example 4:
Can you please give me that item?

## Preferred Output for example 2-4:
Now give all the preferred outputs for the batch of 3 examples.

### Which is best for example 2, Output (a) or Output (b)?:
Output (b)

User explanation: Both outputs are wrong but Output (b) is preferred. Indeed, the real answer is 18.5 years old, which is closer to Output (b) than Output (a).

### Which is best for example 3, Output (a) or Output (b)?:
Output (b)

User explanation: Output (b) is better because Output (a) adds to the to do list "Shoot some hole in the walls", which is not related to tidying a room

### Which is best for example 4, Output (a) or Output (b)?:
Output (a)

User explanation: Output (a) is better because Output (b) adds the sentence "pending verification", which is unnecessary and unrelated to the instruction.

## Annotation starts below
Now is your turn. I will give you a batch of 5 examples.
You should read all the examples first and then select the preferred answers by saying only Output (a) or Output (b) as formatted above without explanation.

## Example 5

### Instruction 5:
{instruction}

### Input 5:
{input}

### Output (a) for example 5:
{output_1}

### Output (b) for example 5:
{output_2}

## Example 6

### Instruction 6:
{instruction}

### Input 6:
{input}

### Output (a) for example 6:
{output_1}

### Output (b) for example 6:
{output_2}

## Example 7

### Instruction 7:
{instruction}

### Input 7:
{input}

### Output (a) for example 7:
{output_1}

### Output (b) for example 7:
{output_2}

## Example 8

### Instruction 8:
{instruction}

### Input 8:
{input}

### Output (a) for example 8:
{output_1}

### Output (b) for example 8:
{output_2}

## Example 9

### Instruction 9:
{instruction}

### Input 9:
{input}

### Output (a) for example 9:
{output_1}

### Output (b) for example 9:
{output_2}

## Preferred Output for example 5-9:
Now give all the preferred outputs for the batch of 5 examples.
<|im_end|>
\end{spverbatim}

\subsection{Multi-turn Chat Dialogue Prompt}
\begin{spverbatim}
    <|im_start|>system
You are helpful assistant whose goal is to decide on a preferred answer to a question in a dialogue.
Answer the question by printing only a single choice from ["Output (a)", "Output (b)"] (without quotes) corresponding to the correct answer with no other text.
<|im_end|>
<|im_start|>user
In this task, we will ask you to select the preferred output AI model's responses.

You will read a batch of examples, which are composed of the following:

1. A beginning of a conversation between an AI system ("Assistant") and a user ("User")
3. Output (a), the first output from the AI system
4. Output (b), the first output from the AI system

I will give you a batch of 5 examples. Your task is to decide which response is better for each example. You should answer using only Output (a) or Output (b) depending on which response is better.

You should read all the examples first and then select the preferred answers by saying only Output (a) or Output (b) as formatted above without explanation.

## Example 1

### Instruction 1:
{instruction}

### Output (a) for example 1:
{output_1}

### Output (b) for example 1:
{output_2}

## Example 2

### Instruction 2:
{instruction}

### Output (a) for example 2:
{output_1}

### Output (b) for example 2:
{output_2}

## Example 3

### Instruction 3:
{instruction}

### Output (a) for example 3:
{output_1}

### Output (b) for example 3:
{output_2}

## Example 4

### Instruction 4:
{instruction}

### Output (a) for example 4:
{output_1}

### Output (b) for example 4:
{output_2}

## Example 5

### Instruction 5:
{instruction}

### Output (a) for example 5:
{output_1}

### Output (b) for example 5:
{output_2}

## Preferred Output for example 1-5:
Now give all the preferred outputs for the batch of 5 examples.
<|im_end|>
\end{spverbatim}

\end{document}